\documentclass[lettersize,journal]{IEEEtran}
\usepackage{amsmath,amsfonts}
\usepackage{algorithmic}
\usepackage{array}
\usepackage{needspace}
\usepackage[caption=false,font=normalsize,labelfont=sf,textfont=sf]{subfig}
\usepackage{textcomp}
\usepackage{stfloats}
\usepackage{url}
\usepackage{verbatim}
\usepackage{graphicx}
\usepackage{enumitem}
\usepackage{tabularx}
\usepackage{booktabs,makecell,multirow}
\usepackage{xcolor}
\usepackage{balance}
\usepackage{tcolorbox}
\tcbuselibrary{skins}
\usepackage{cite}
\usepackage[colorlinks=true, citecolor=blue, linkcolor=black, urlcolor=blue]{hyperref}
\def\BibTeX{{\rm B\kern-.05em{\sc i\kern-.025em b}\kern-.08em
    T\kern-.1667em\lower.7ex\hbox{E}\kern-.125emX}}

\begin{document}
\title{In-Context Model Predictive Generation: Open-Vocabulary Motion Synthesis from Language Models to Physics}

\author{
Xiaomeng~Fu,
Junfan~Lin,
Yang~Liu,
Yaowei~Wang,
Guanbin~Li,
Liang~Lin,
Ziliang~Chen

\IEEEcompsocitemizethanks{\IEEEcompsocthanksitem Xiaomeng Fu, Guanbin Li and Liang Lin are with the School of Computer Science and Engineering, Sun Yat-sen University, Guangzhou, China, and are also with the Department of Networked Intelligence, Peng Cheng Laboratory, Shenzhen, China (e-mail: fuxm7@mail2.sysu.edu.cn,liguanbin@mail.sysu.edu.cn,linliang@ieee.org).
\IEEEcompsocthanksitem Yang Liu is with the School of Computer Science and Engineering, Sun Yat-sen University, Guangzhou, China (e-mail: liuy856@mail.sysu.edu.cn).  
\IEEEcompsocthanksitem Yaowei Wang is with the School of Computer Science and Engineering, Harbin Institute of Technology, Shenzhen, and the Department of Networked Intelligence, Peng Cheng Laboratory, Shenzhen, China (e-mail: wangyaowei@hit.edu.cn).
\IEEEcompsocthanksitem Junfan Lin and Ziliang Chen are with the Department of Networked Intelligence, Peng Cheng Laboratory, Shenzhen, China (e-mail:linjf8@mail2.sysu.edu.cn, c.ziliang@yahoo.com).
\IEEEcompsocthanksitem  Corresponding authors are Guanbin Li and Ziliang Chen.
}
}

\maketitle

\begin{abstract}
Synthesizing human motion from textual descriptions is essential for immersive digital applications, yet existing methods face a persistent trade-off between semantic fidelity and physical realism. Large language model (LLM)-based approaches can interpret diverse open-vocabulary instructions and compose high-level action plans, but they often generate motions that violate physical constraints. Physics-aware models improve realism through simulation or control, but they struggle with semantic complexity, fine-grained instructions, and novel concepts. To address this gap, we propose In-Context Model Predictive Generation (ICMPG), a framework that integrates language-model planning with inference-time physical feedback. ICMPG reformulates motion synthesis as a Model Predictive Control (MPC)-like process with two modules. The Context-Aware Motion Generation (CAMG) module uses an LLM as a planner to decompose textual commands and generate candidate motion sequences from motion tokens. The Model Predictive Generation (MPG) module evaluates these candidates through physical simulation and semantic alignment, estimates a composite reward, and selects the best sequence to guide subsequent generation steps. Unlike open-loop generation, this closed-loop refinement enables ICMPG to adapt motions to both the input semantics and the simulated physical environment without task-specific policy retraining. Extensive experiments across standard and zero-shot open-vocabulary settings show that ICMPG generalizes robustly to diverse commands and produces motions that are more physically plausible and semantically faithful than representative baselines on the evaluated benchmarks. The framework bridges semantic interpretation and physical simulation while remaining flexible enough to incorporate different LLM backbones, enabling more versatile and controllable text-driven motion synthesis.
\end{abstract}


\newcommand{\figExample}{
    \begin{figure}[!t]
        \centering
        \includegraphics[width=0.6\textwidth]{figures/example.png}
        \caption{Example Figure.}
        \label{fig:example}
    \end{figure}
}

\newcommand{\figmethod}{
    \begin{figure*}[!t]
        \centering
        \includegraphics[width=\linewidth]{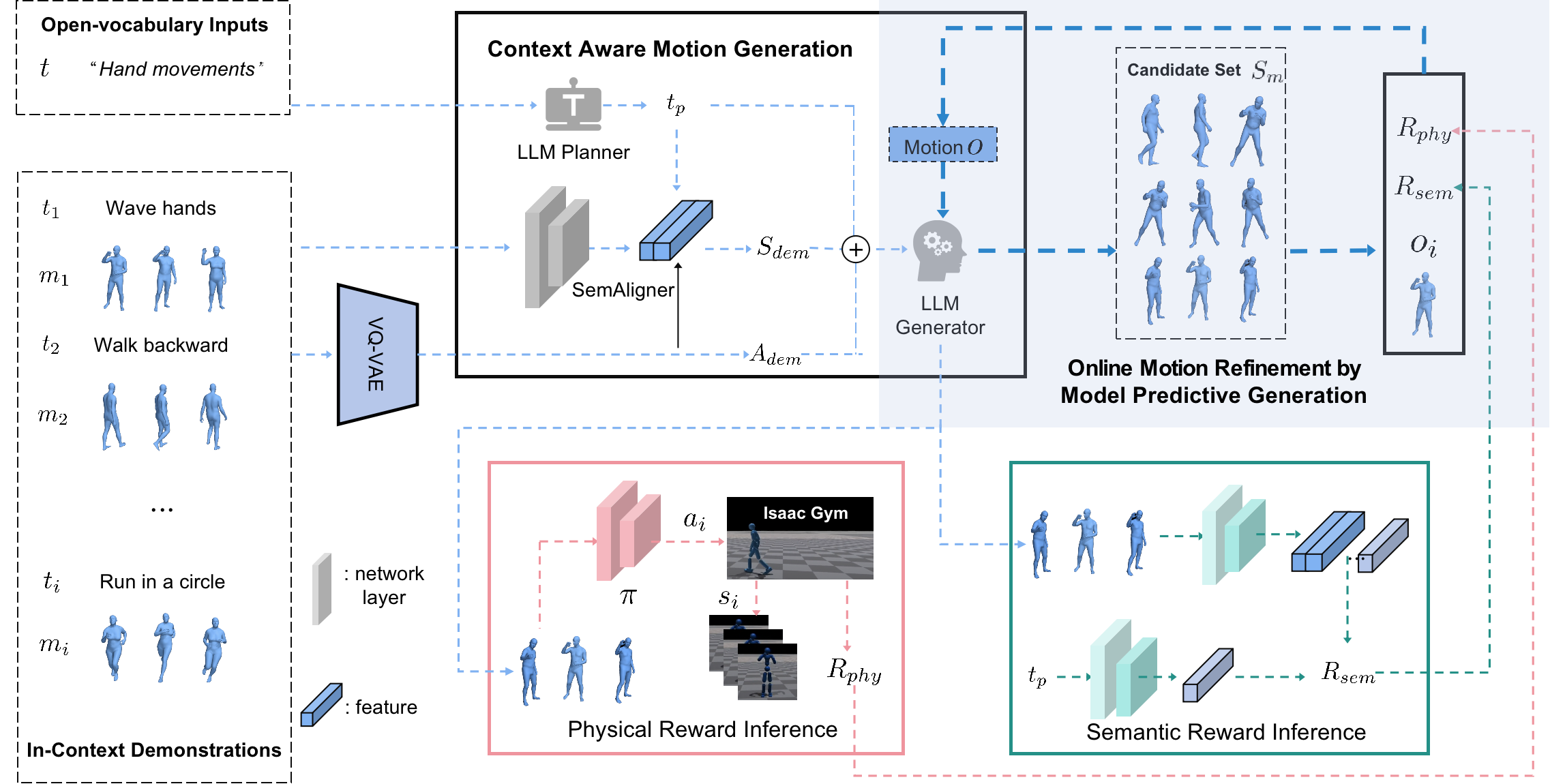}
        \caption{Complete Workflow of the In-Context Model Predictive Generation (ICMPG) Framework. The pipeline begins with a one-time offline preparation phase, where a VQ-VAE, SemAligner, and a pre-trained policy $\pi$ are initialized for motion encoding, demonstration retrieval, and physical simulation, respectively. In the online motion refinement phase, the LLM Planner decomposes the input text $t$ into a structured target prompt $t_p$ using atomic demonstrations $A_{dem}$. The LLM Generator then combines $t_p$ with semantic demonstrations $S_{dem}$ to produce candidate motion sequences $S_m$. The Model Predictive Generation (MPG) module evaluates these candidates using the physical reward $\mathcal{R}_{phy}$ and semantic reward $\mathcal{R}_{sem}$, selects the optimal motion sequence $o_i$, and appends it to the history $O$ for the next generation step.}
        \label{fig:over}
    \end{figure*}
}

\newcommand{\figcontrast}{
    \begin{figure*}[!t]
        \centering
        \begin{minipage}{0.15\textwidth}
            \textbf{MotionCLIP} \\
        \end{minipage}%
        \begin{minipage}{0.4\textwidth}
            \includegraphics[width=\linewidth, height=2.2cm]{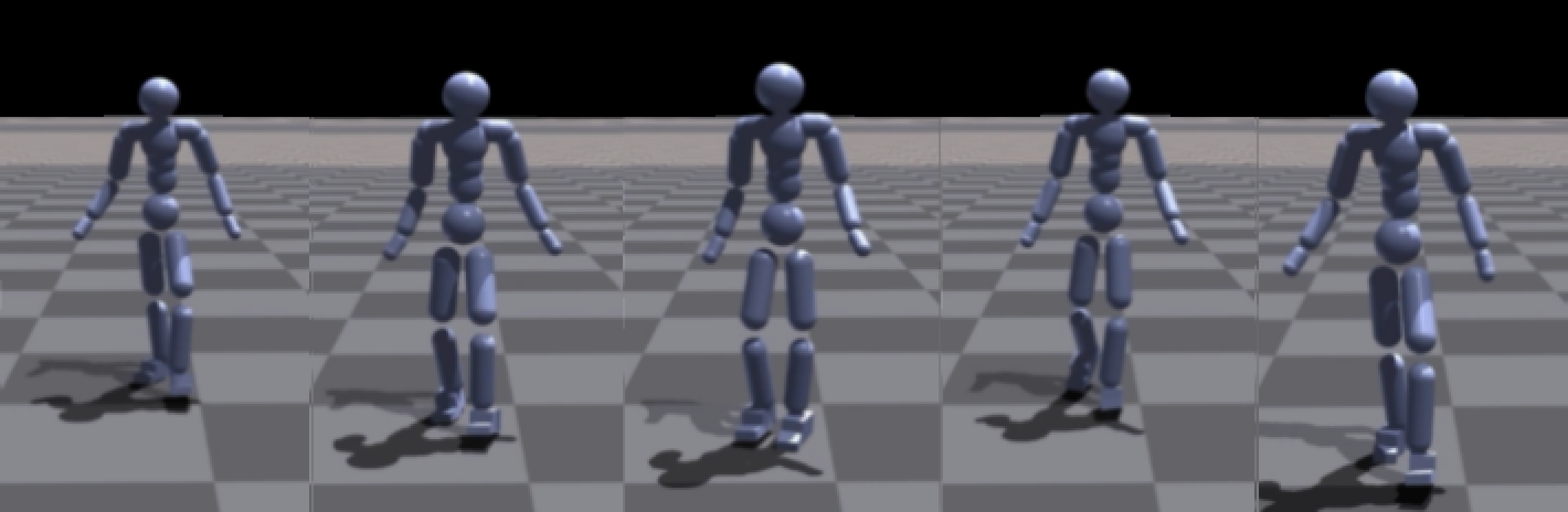}
        \end{minipage}
        \begin{minipage}{0.4\textwidth}
            \includegraphics[width=\linewidth, height=2.2cm]{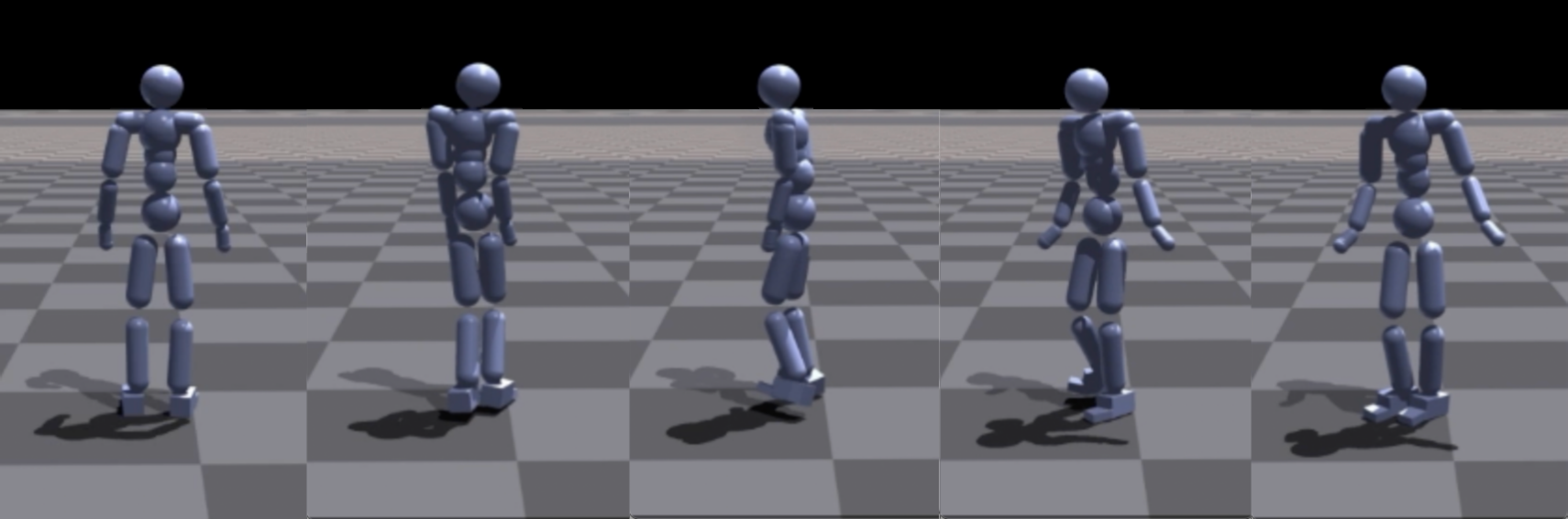}
        \end{minipage}
        \vspace{0.1cm}

        \begin{minipage}{0.15\textwidth}
            \textbf{AnySkill} \\
        \end{minipage}%
        \begin{minipage}{0.4\textwidth}
            \includegraphics[width=\linewidth, height=2.2cm]{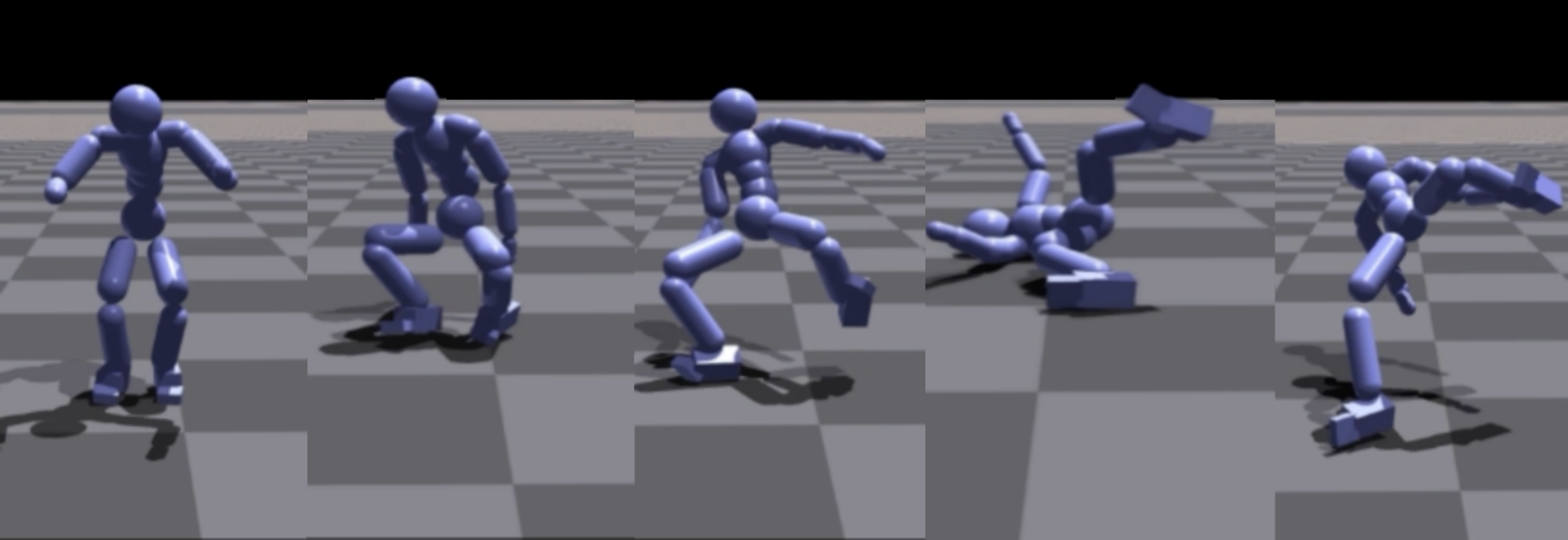}
        \end{minipage}
        \begin{minipage}{0.4\textwidth}
            \includegraphics[width=\linewidth, height=2.2cm]{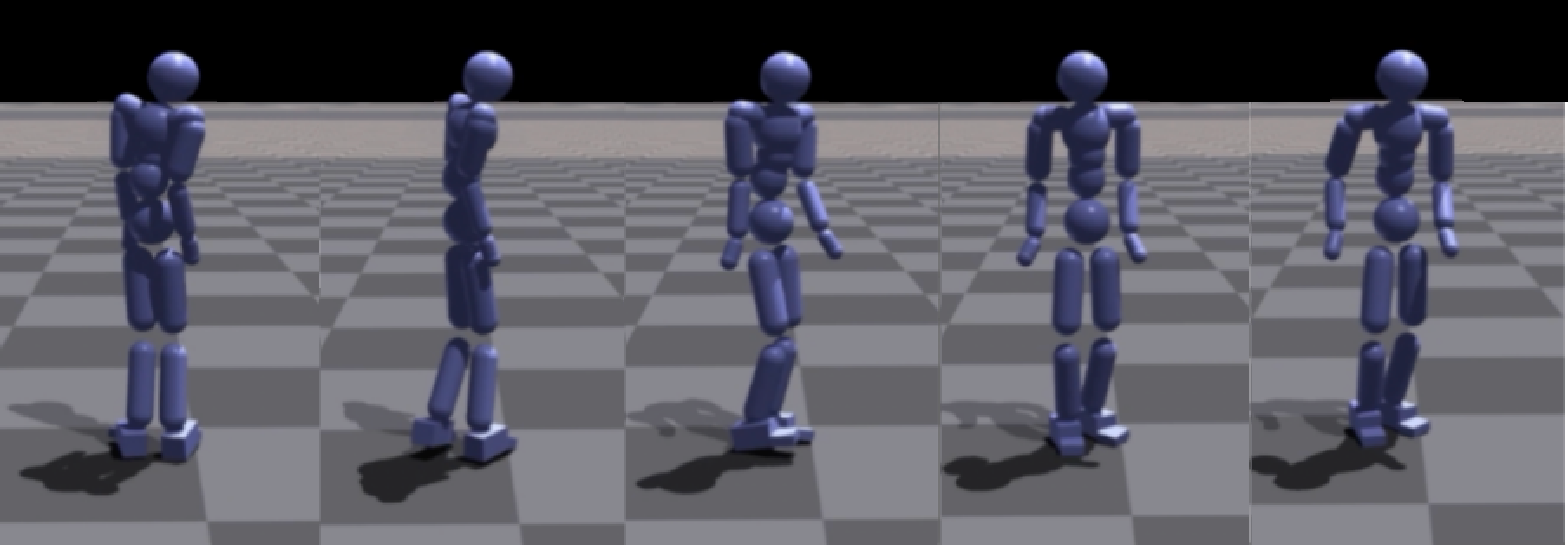}
        \end{minipage}
        \vspace{0.1cm}

        \begin{minipage}{0.145\textwidth}
            \textbf{ICMPG} \\
        \end{minipage}%
        \begin{minipage}{0.4\textwidth}
            \includegraphics[width=\linewidth, height=2.2cm]{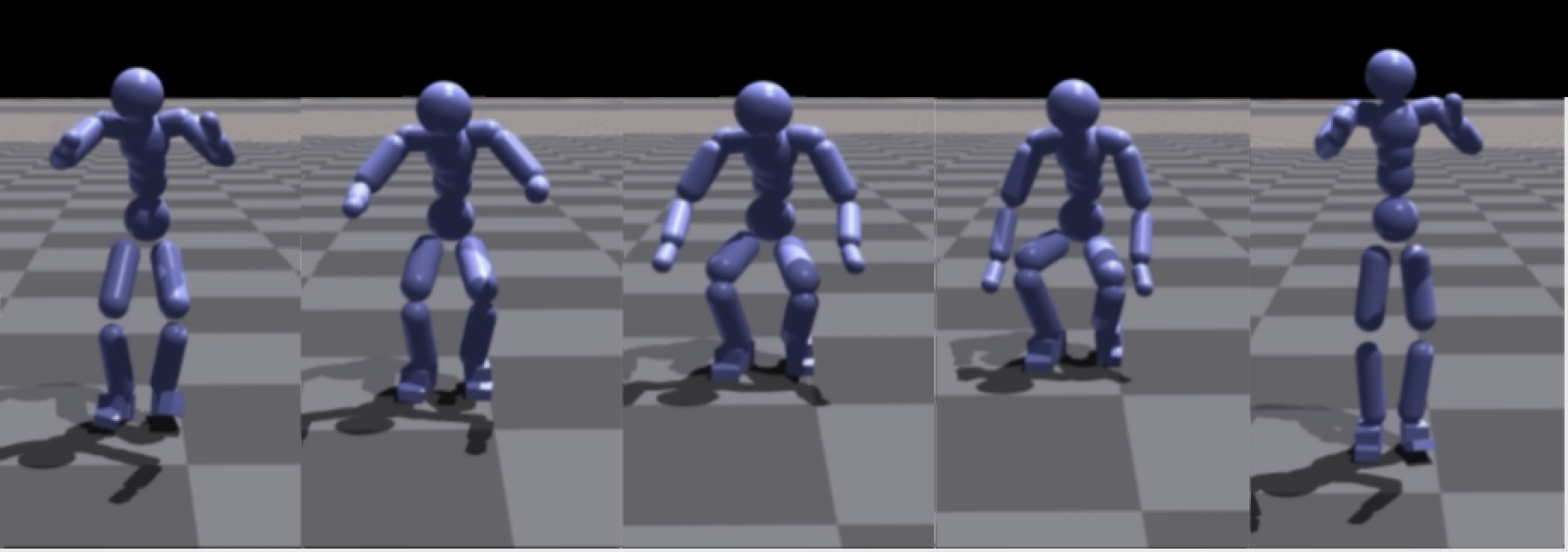}
        \end{minipage}
        \begin{minipage}{0.4\textwidth}
            \includegraphics[width=\linewidth, height=2.2cm]{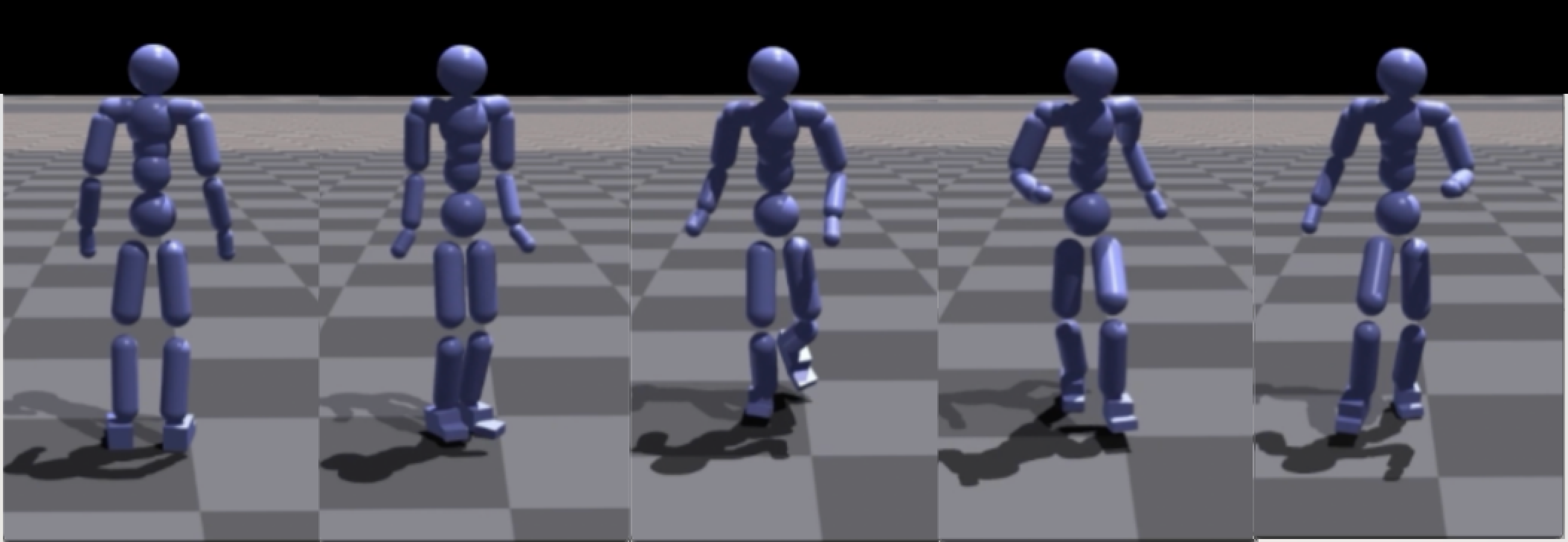}
        \end{minipage}
        \caption{Comparison of Motions Under Physical Simulation. We compare ICMPG with MotionCLIP and AnySkill using the input texts ``Perform a squat'' and ``Sprint backwards.'' ICMPG generates motions that align with the input semantics and exhibit strong physical stability during simulation.}
        \label{fig:physs}
    \end{figure*}
}

\newcommand{\figopenvocab}{
    \begin{figure*}[!t]
        \centering

        \begin{minipage}{0.15\textwidth}
            \textbf{MotionCLIP} \\
        \end{minipage}%
        \begin{minipage}{0.85\textwidth}
            \begin{minipage}{0.48\linewidth}
                \includegraphics[width=\linewidth]{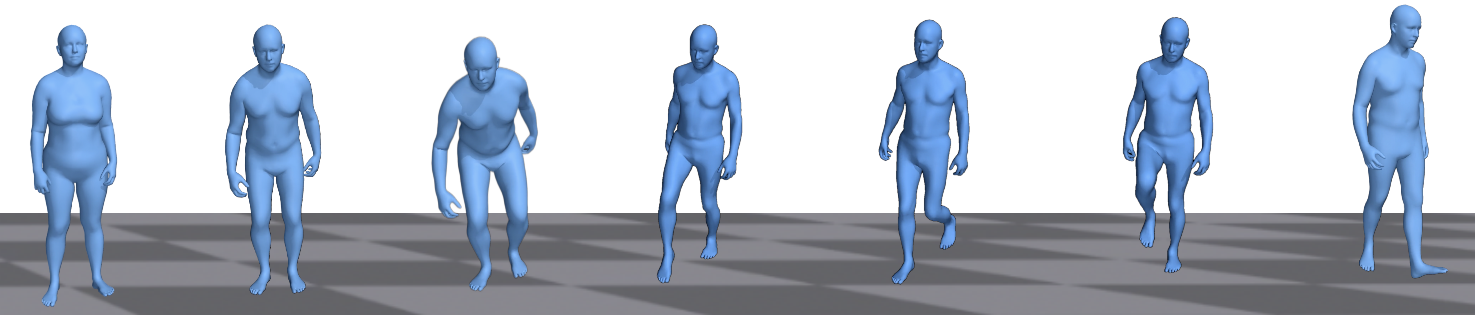}
            \end{minipage}%
            \hfill
            \begin{minipage}{0.48\linewidth}
                \includegraphics[width=\linewidth]{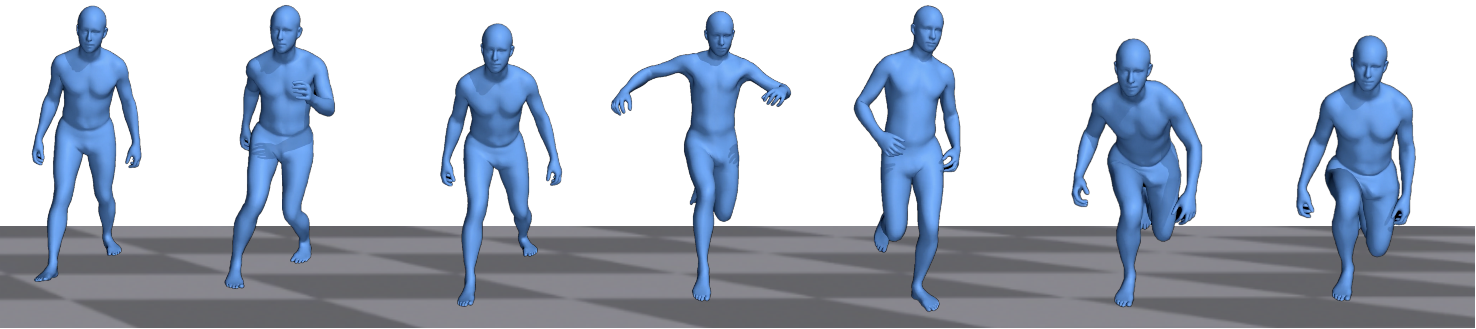}
            \end{minipage}
        \end{minipage}

        \vspace{0.5cm}

        \begin{minipage}{0.15\textwidth}
            \textbf{AnySkill} \\
        \end{minipage}%
        \begin{minipage}{0.85\textwidth}
            \begin{minipage}{0.48\linewidth}
                \includegraphics[width=\linewidth]{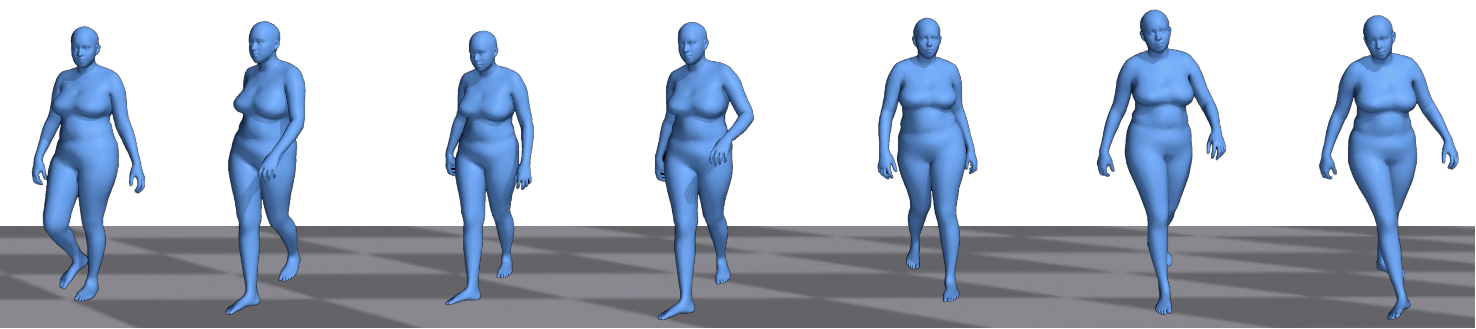}
            \end{minipage}%
            \hfill
            \begin{minipage}{0.48\linewidth}
                \includegraphics[width=\linewidth]{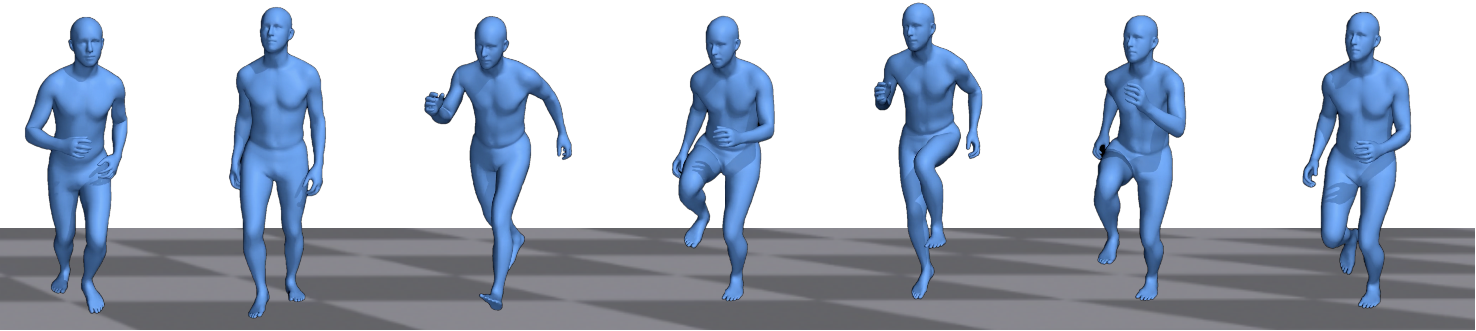}
            \end{minipage}
        \end{minipage}

        \vspace{0.5cm}

        \begin{minipage}{0.15\textwidth}
            \textbf{ICMPG} \\
        \end{minipage}%
        \begin{minipage}{0.85\textwidth}
            \begin{minipage}{0.48\linewidth}
                \includegraphics[width=\linewidth]{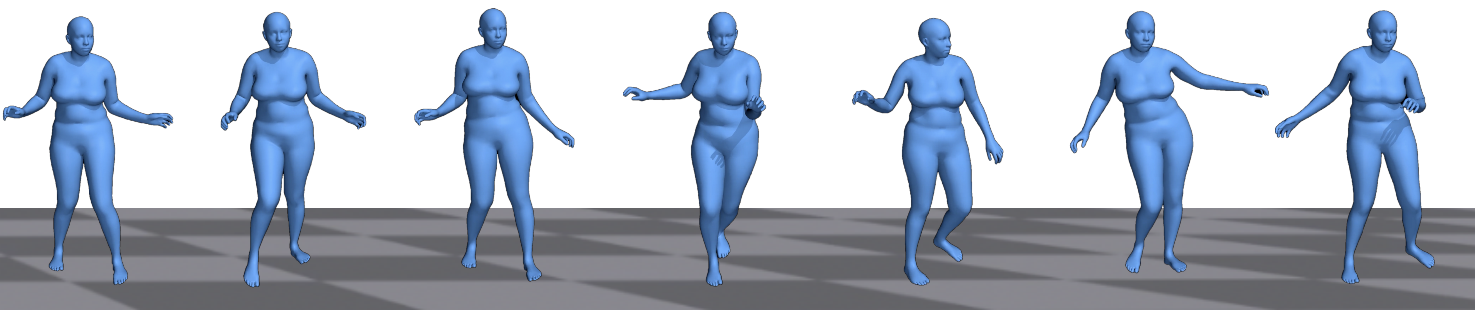}
            \end{minipage}%
            \hfill
            \begin{minipage}{0.48\linewidth}
                \includegraphics[width=\linewidth]{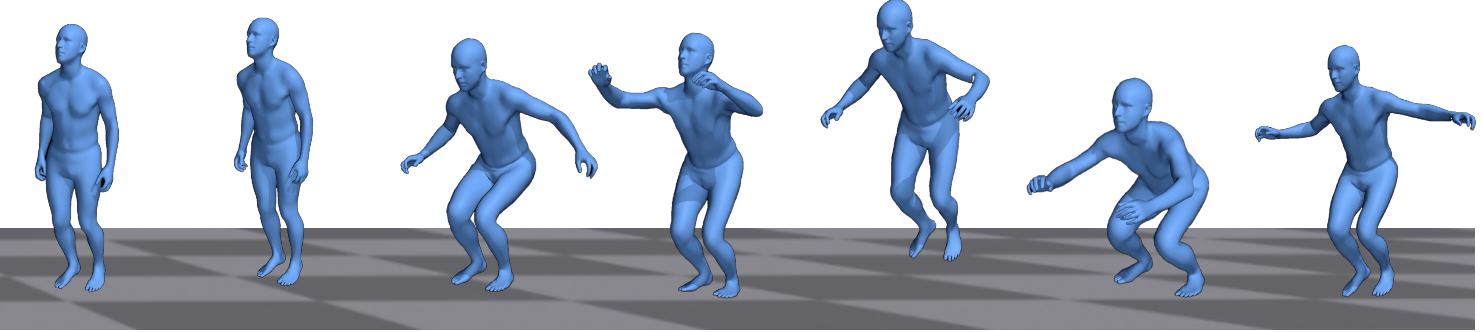}
            \end{minipage}
        \end{minipage}

        \caption{Comparison Under Open-Vocabulary Textual Descriptions. We compare ICMPG with MotionCLIP and AnySkill using the input texts ``Ballroom dance in pattern series'' and ``Hands up high jump.'' ICMPG generates motions that better align with both textual descriptions.}
        \label{fig:qualitative_results}
    \end{figure*}
}

\newcommand{\figlong}{
    \begin{figure*}[!t]
        \centering
        \subfloat[``Squat in frog pose, stand up.'']{
            \includegraphics[width=0.45\textwidth]{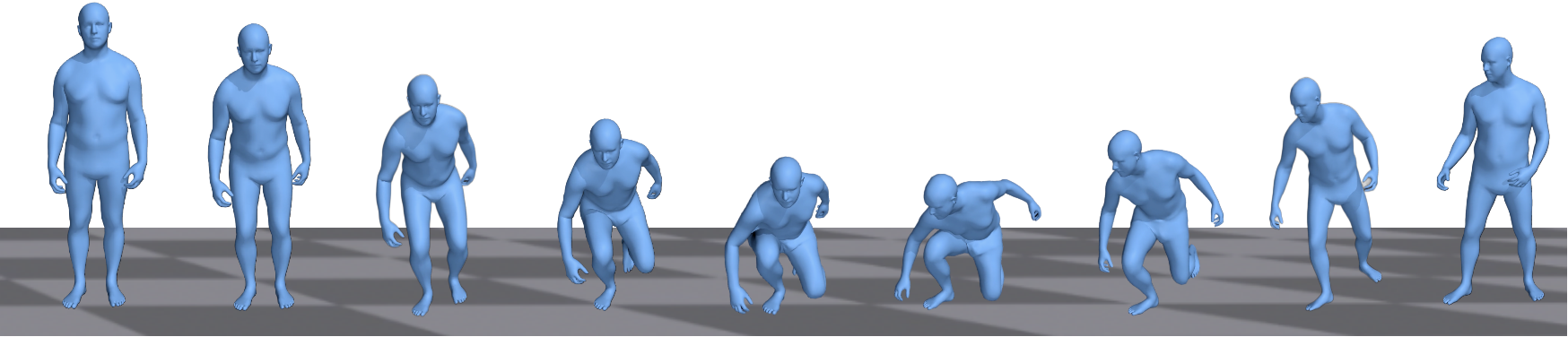}
            \label{fig:psubA}
        }
        \hfill
        \subfloat[``Pick something up, lift.'']{
            \includegraphics[width=0.45\textwidth]{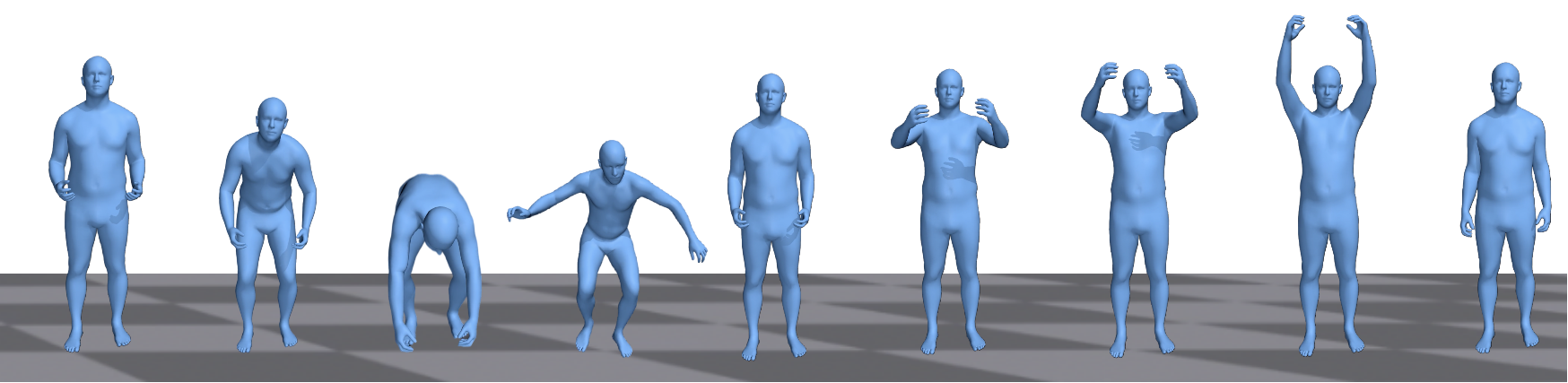}
            \label{fig:psubB}
        }
        \\
        \subfloat[``Hand stretches, leg stretches.'']{
            \includegraphics[width=0.45\textwidth]{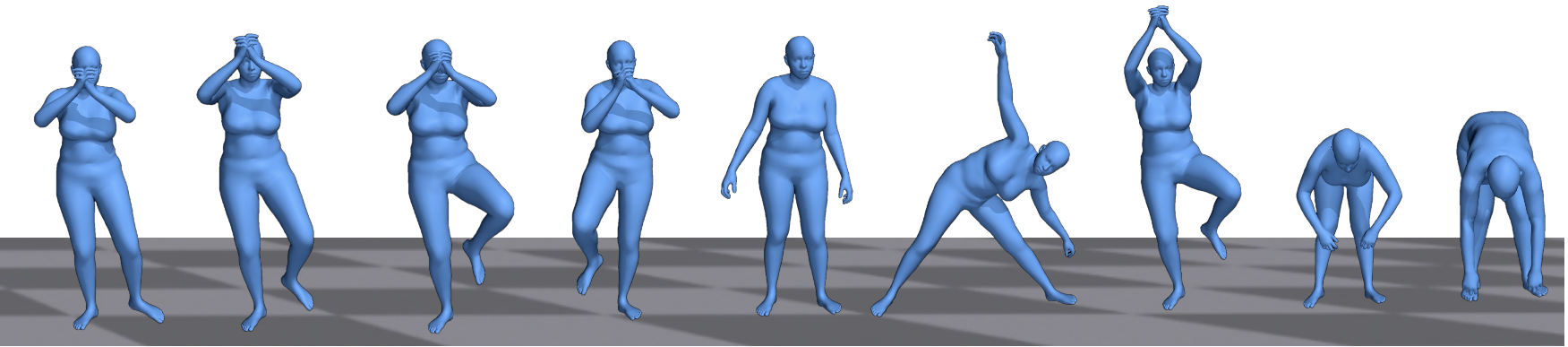}
            \label{fig:psubC}
        }
        \hfill
        \subfloat[``Walking casually, single-leg standing.'']{
            \includegraphics[width=0.45\textwidth]{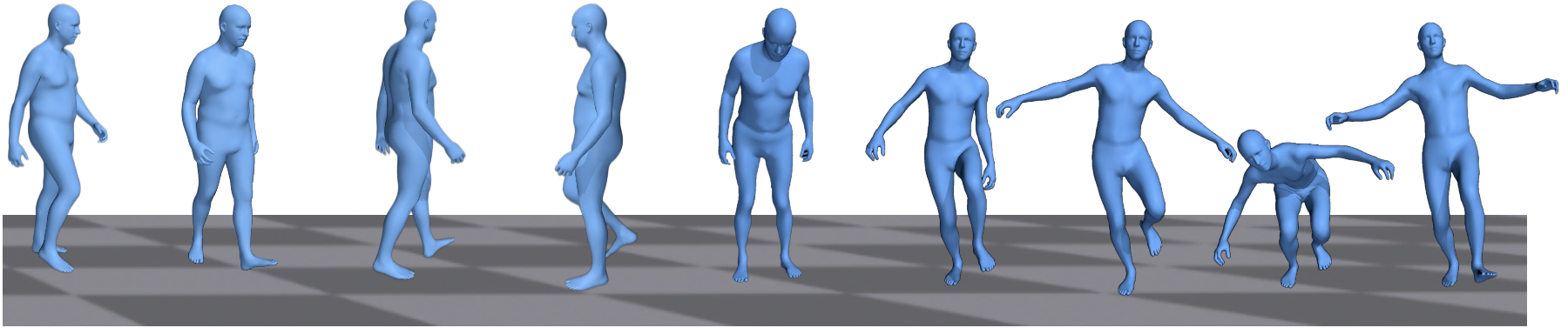}
            \label{fig:psubD}
        }
        \caption{Long-Sequence Motion Generation Examples Using the BABEL Annotation Protocol. The generated motions exhibit semantic alignment with the input texts and illustrate semantic progression between motion segments.}
        \label{figlong}
    \end{figure*}
}

\newcommand{\figintro}{
    \begin{figure}[!t]
        \centering
        \begin{minipage}{0.42\textwidth}
            \centering
            \includegraphics[width=\textwidth]{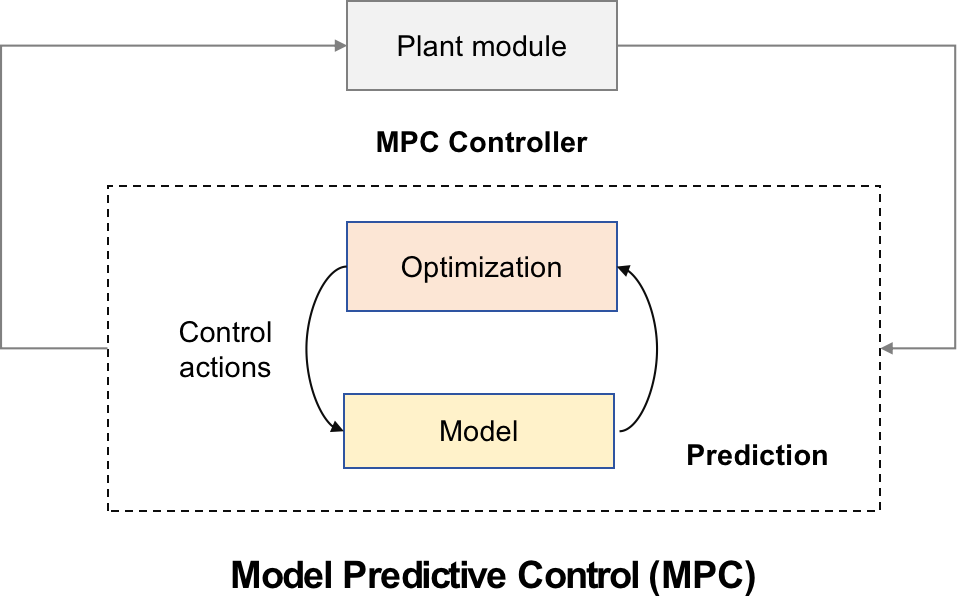}
        \end{minipage}

        \vspace{0.5em}

        \begin{minipage}{0.42\textwidth}
            \centering
            \includegraphics[width=\textwidth]{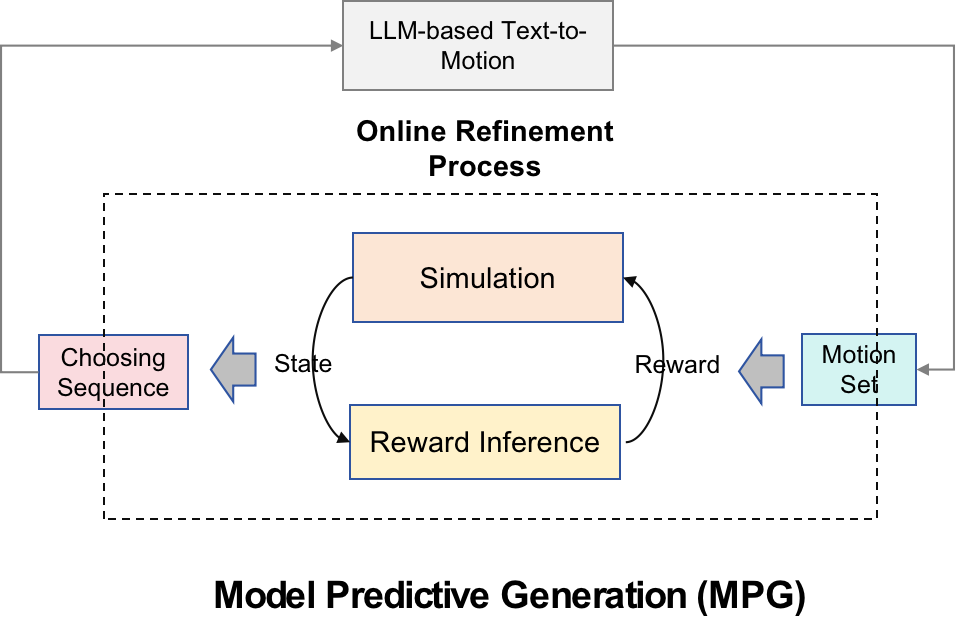}
        \end{minipage}
        \caption{Conceptual Analogy Between Classical Model Predictive Control (MPC) and the Proposed In-Context Model Predictive Generation (ICMPG) Framework. Top: a traditional MPC controller optimizes low-level control actions to govern a physical plant's state. Bottom: ICMPG adapts this paradigm to open-vocabulary motion synthesis, where the LLM Generator acts as the predictive model and the control action corresponds to selecting the optimal motion token according to physical and semantic rewards.}
        \label{fig:intro}
    \end{figure}
}

\newcommand{\figablation}{
    \begin{figure}[!t]
        \centering
        \includegraphics[width=0.45\textwidth, height=3.6cm]{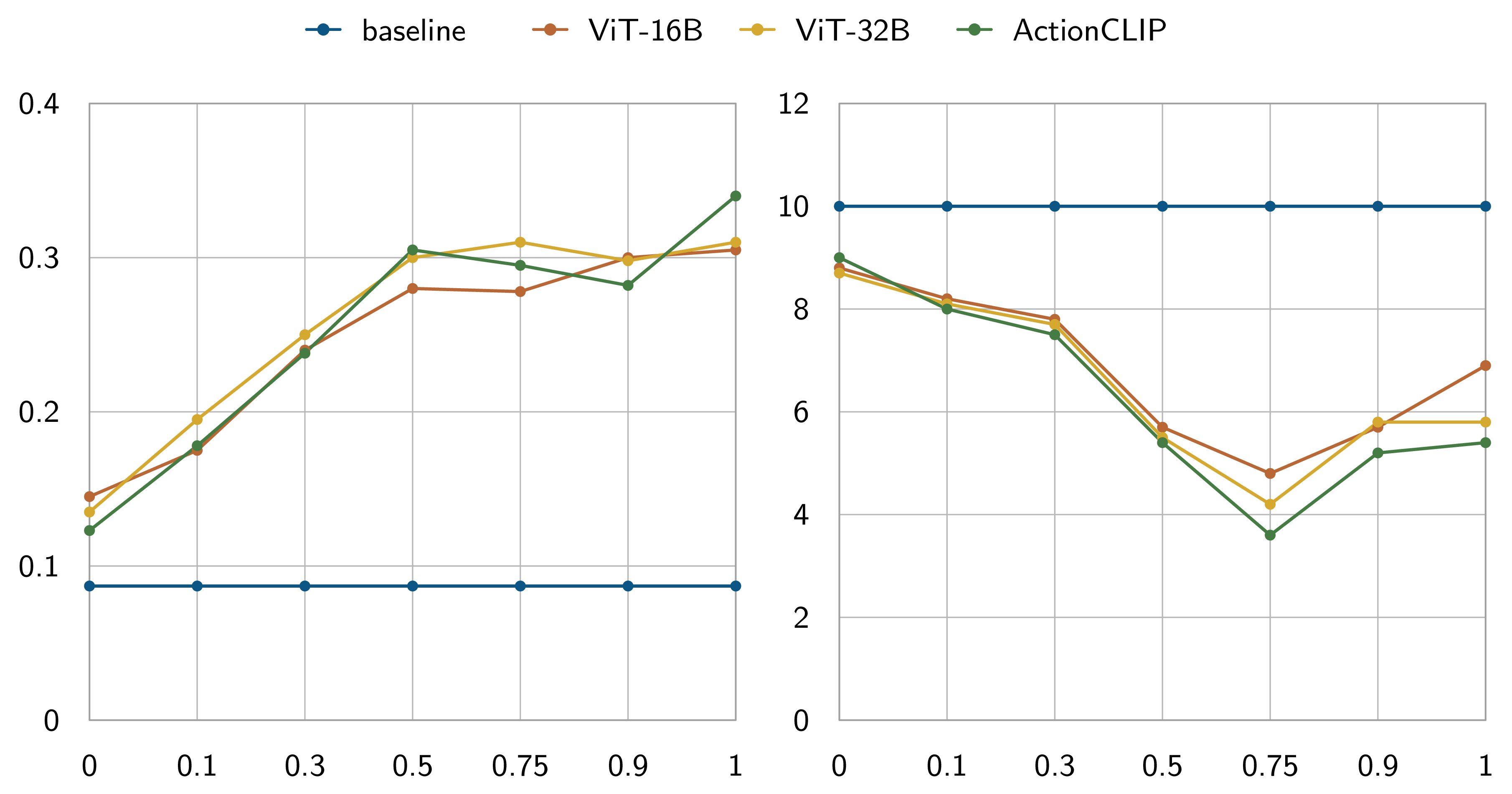}
        \caption{Evaluation of the Reward Function. We conduct an ablation study to evaluate the impact of the hyperparameter $\beta$ on model performance, as well as the effect of employing different semantic models during the semantic reward inference stage. Experimental results show that, regardless of the semantic model used, the performance trend of ICMPG with respect to $\beta$ remains consistent. The hyperparameter $\beta$ effectively governs the trade-off between semantic consistency and physical plausibility in the generated motion sequences. We select \(\beta=1\) to prioritize semantic alignment while retaining acceptable physical error; \(\beta=0.75\) yields the lowest Phys-Err.}
        \label{fig:m}
    \end{figure}
}

\newcommand{\tabablaLLM}{
    \begin{table}[!t]
    \caption{Impact of LLM Backbones on ICMPG Performance. We compare ICMPG variants based on LLaMA-2, LLaMA-3.1, and Phi-4. The results indicate that the reasoning capabilities of LLMs play an important role in motion generation quality.}
    \label{tab:LLM}
    \centering
    \begin{tabular}{c|c|c}
        \toprule
        Method & CLIP\_Score $\uparrow$ & Phys-Err $\downarrow$ \\
        \midrule
        Real & 41.94 & 1.637 \\
        \midrule
        Ours (LLaMA-2) & 16.94 & 14.933 \\
        Ours (LLaMA-3.1) & 26.25 & 4.583 \\
        Ours (Phi-4) & 24.47 & 4.782 \\
        \bottomrule
    \end{tabular}
    \end{table}
}

\newcommand{\tabopen}{
    \setlength{\tabcolsep}{10pt}
    \begin{table*}[!t]
    \caption{Comparison on Open-Vocabulary Motion Generation Tasks. We evaluate ICMPG and representative open-vocabulary motion generation methods on the BABEL dataset. The LoRA-finetuned variant achieves the best semantic metrics among the compared methods while maintaining competitive physical plausibility.}
    \label{tab:com}
    \centering
    \begin{tabular}{c|l|l l l|l l}
        \toprule
        \multirow{2}{*}{Method} & \multirow{2}{*}{FID $\downarrow$} & \multirow{2}{*}{CLIP\_Score $\uparrow$} & \multicolumn{2}{c|}{R-Precision $\uparrow$} & \multirow{2}{*}{Success $\uparrow$} & \multirow{2}{*}{Phys-Err $\downarrow$} \\ \cline{4-5}
        & & & Top-1 & Top-3 & & \\
        \midrule
        Real & 0.002 & 41.94 & 0.409 & 0.716 & 100\% & 1.637 \\
        \midrule
        AvatarCLIP~\cite{hong2022avatarclip} & 0.642 & 23.67 & 0.274 & 0.486 & 51.7\% & 24.162 \\
        MotionCLIP~\cite{tevet2022motionclip} & \textbf{$\textbf{0.630}^{\mathbf{3}}$} & 24.19 & 0.266 & 0.521 & 44.3\% & 31.433 \\
        AnySkill~\cite{cui2024anyskill} & 0.635 & \textbf{$\textbf{28.64}^{\mathbf{2}}$} & \textbf{$\textbf{0.291}^{\mathbf{2}}$} & \textbf{$\textbf{0.568}^{\mathbf{2}}$} & \textbf{$\textbf{84.1\%}^{\mathbf{1}}$} & \textbf{$\textbf{3.675}^{\mathbf{1}}$} \\
        \midrule
        Ours & \textbf{$\textbf{0.589}^{\mathbf{2}}$} & \textbf{$\textbf{26.25}^{\mathbf{3}}$} & \textbf{$\textbf{0.279}^{\mathbf{3}}$} & \textbf{$\textbf{0.554}^{\mathbf{3}}$} & \textbf{$\textbf{77.2\%}^{\mathbf{3}}$} & \textbf{$\textbf{4.583}^{\mathbf{3}}$} \\
        Ours (LoRA) & \textbf{$\textbf{0.492}^{\mathbf{1}}$} & \textbf{$\textbf{29.32}^{\mathbf{1}}$} & \textbf{$\textbf{0.316}^{\mathbf{1}}$} & \textbf{$\textbf{0.603}^{\mathbf{1}}$} & \textbf{$\textbf{82.9\%}^{\mathbf{2}}$} & \textbf{$\textbf{3.704}^{\mathbf{2}}$} \\
        \bottomrule
    \end{tabular}
    \end{table*}
}

\newcommand{\tabnormal}{
    \begin{table*}[!t]
        \caption{Comparison of Text-to-Motion Generation Methods on KIT-ML. We evaluate ICMPG and several leading text-driven human motion generation models in terms of motion realism and semantic consistency.}
        \label{tab:com1}
        \centering
        \begin{tabular}{l|l|l|l|l|l|l|l|l}
            \toprule
            \multirow{2}{*}{Method} &
            \multirow{2}{*}{FID $\downarrow$} &
            \multicolumn{2}{c|}{R-Precision $\uparrow$} &
            \multirow{2}{*}{Diversity $\rightarrow$} &
            \multirow{2}{*}{Penetrate $\downarrow$} &
            \multirow{2}{*}{Float $\downarrow$} &
            \multirow{2}{*}{Skate $\downarrow$} &
            \multirow{2}{*}{Phys-Err $\downarrow$} \\
            \cline{3-4}
            & & Top-1 & Top-3 & & & & & \\
            \midrule
            Real & 0.031 & 0.424 & 0.779 & 11.08 & 0.145 & 0.229 & 0.060 & 0.434 \\
            Ours VQ-VAE & 0.438 & 0.412 & 0.757 & 9.997 & 1.176 & 2.338 & 1.449 & 4.963 \\
            \midrule
            MDM~\cite{tevet2022human} & \textbf{$\textbf{0.497}^{\mathbf{1}}$} & 0.221 & 0.396 & 10.847 & 13.384 & 19.342 & \textbf{$\textbf{1.879}^{\mathbf{2}}$} & 34.605 \\
            MotionDiffuse~\cite{zhang2022motiondiffuse} & 1.954 & \textbf{$\textbf{0.417}^{\mathbf{1}}$} & 0.739 & \textbf{$\textbf{11.100}^{\mathbf{1}}$} & 23.973 & \textbf{$\textbf{6.988}^{\mathbf{3}}$} & 3.925 & 34.886 \\
            T2M-GPT~\cite{zhang2023generating} & \textbf{$\textbf{0.514}^{\mathbf{2}}$} & \textbf{$\textbf{0.416}^{\mathbf{2}}$} & \textbf{$\textbf{0.745}^{\mathbf{2}}$} & 10.921 & 12.482 & 7.683 & 3.104 & 23.269 \\
            MotionGPT~\cite{jiang_motiongpt:_2023} & 0.737 & 0.414 & 0.731 & \textbf{$\textbf{10.974}^{\mathbf{2}}$} & 11.478 & 8.023 & 3.836 & 23.337 \\
            AttT2M~\cite{zhong2023attt2m} & 0.870 & 0.413 & \textbf{$\textbf{0.751}^{\mathbf{1}}$} & \textbf{$\textbf{10.960}^{\mathbf{3}}$} & \textbf{$\textbf{10.477}^{\mathbf{3}}$} & 7.624 & 3.571 & \textbf{$\textbf{21.672}^{\mathbf{3}}$} \\
            \midrule
            Ours & 0.771 & 0.399 & 0.727 & 10.525 & \textbf{$\textbf{1.328}^{\mathbf{2}}$} & \textbf{$\textbf{2.487}^{\mathbf{2}}$} & \textbf{$\textbf{1.982}^{\mathbf{3}}$} & \textbf{$\textbf{5.797}^{\mathbf{2}}$} \\
            Ours (LoRA) & \textbf{$\textbf{0.635}^{\mathbf{3}}$} & \textbf{$\textbf{0.415}^{\mathbf{3}}$} & \textbf{$\textbf{0.741}^{\mathbf{3}}$} & 10.872 & \textbf{$\textbf{0.948}^{\mathbf{1}}$} & \textbf{$\textbf{2.439}^{\mathbf{1}}$} & \textbf{$\textbf{1.751}^{\mathbf{1}}$} & \textbf{$\textbf{5.138}^{\mathbf{1}}$} \\
            \bottomrule
        \end{tabular}
    \end{table*}
}

\newcommand{\tabnormall}{
    \begin{table*}[!t]
        \caption{Comparison of Text-to-Motion Generation Methods on HumanML3D. We evaluate ICMPG and several leading text-driven human motion generation models in terms of motion realism and semantic consistency.}
        \label{tab:com2}
        \centering
        \begin{tabular}{l|l|l|l|l|l|l|l|l}
            \toprule
            \multirow{2}{*}{Method} &
            \multirow{2}{*}{FID $\downarrow$} &
            \multicolumn{2}{c|}{R-Precision $\uparrow$} &
            \multirow{2}{*}{Diversity $\rightarrow$} &
            \multirow{2}{*}{Penetrate $\downarrow$} &
            \multirow{2}{*}{Float $\downarrow$} &
            \multirow{2}{*}{Skate $\downarrow$} &
            \multirow{2}{*}{Phys-Err $\downarrow$} \\
            \cline{3-4}
            & & Top-1 & Top-3 & & & & & \\
            \midrule
            Real & 0.002 & 0.511 & 0.797 & 9.503 & 0.131 & 0.327 & 0.047 & 0.505 \\
            Ours VQ-VAE & 0.091 & 0.501 & 0.787 & 9.597 & 0.136 & 0.338 & 0.049 & 0.523 \\
            \midrule
            MDM~\cite{tevet2022human} & 0.544 & 0.320 & 0.611 & \textbf{$\textbf{9.561}^{\mathbf{2}}$} & 11.291 & 18.876 & 1.406 & 31.572 \\
            MotionDiffuse~\cite{zhang2022motiondiffuse} & 0.630 & \textbf{$\textbf{0.491}^{\mathbf{3}}$} & \textbf{$\textbf{0.782}^{\mathbf{2}}$} & 9.410 & 20.278 & 6.450 & 3.925 & 30.652 \\
            T2M-GPT~\cite{zhang2023generating} & \textbf{$\textbf{0.116}^{\mathbf{2}}$} & \textbf{$\textbf{0.491}^{\mathbf{3}}$} & 0.775 & 9.722 & 10.908 & 7.649 & 2.341 & 20.898 \\
            MotionGPT~\cite{jiang_motiongpt:_2023} & \textbf{$\textbf{0.234}^{\mathbf{3}}$} & \textbf{$\textbf{0.492}^{\mathbf{2}}$} & 0.778 & \textbf{$\textbf{9.528}^{\mathbf{1}}$} & 11.774 & 7.125 & 2.459 & 21.358 \\
            AttT2M~\cite{zhong2023attt2m} & \textbf{$\textbf{0.112}^{\mathbf{1}}$} & \textbf{$\textbf{0.499}^{\mathbf{1}}$} & \textbf{$\textbf{0.786}^{\mathbf{1}}$} & 9.700 & 10.856 & 7.835 & 2.371 & 21.062 \\
            PhysDiff~\cite{yuan2023physdiff} & 0.551 & 0.487 & \textbf{$\textbf{0.780}^{\mathbf{3}}$} & \textbf{$\textbf{9.421}^{\mathbf{3}}$} & \textbf{$\textbf{0.898}^{\mathbf{1}}$} & \textbf{$\textbf{1.368}^{\mathbf{3}}$} & \textbf{$\textbf{0.423}^{\mathbf{1}}$} & \textbf{$\textbf{2.690}^{\mathbf{3}}$} \\
            \midrule
            Ours & 0.347 & 0.473 & 0.761 & 9.621 & \textbf{$\textbf{1.045}^{\mathbf{3}}$} & \textbf{$\textbf{0.831}^{\mathbf{1}}$} & \textbf{$\textbf{0.774}^{\mathbf{3}}$} & \textbf{$\textbf{2.651}^{\mathbf{2}}$} \\
            Ours (LoRA) & \textbf{$\textbf{0.116}^{\mathbf{2}}$} & \textbf{$\textbf{0.491}^{\mathbf{3}}$} & \textbf{$\textbf{0.780}^{\mathbf{3}}$} & 9.621 & \textbf{$\textbf{1.023}^{\mathbf{2}}$} & \textbf{$\textbf{0.834}^{\mathbf{2}}$} & \textbf{$\textbf{0.765}^{\mathbf{2}}$} & \textbf{$\textbf{2.622}^{\mathbf{1}}$} \\
            \bottomrule
        \end{tabular}
    \end{table*}
}

\newcommand{\figcode}{
    \begin{figure}[!t]
        \centering
        \includegraphics[width=0.44\textwidth, height=3.6cm]{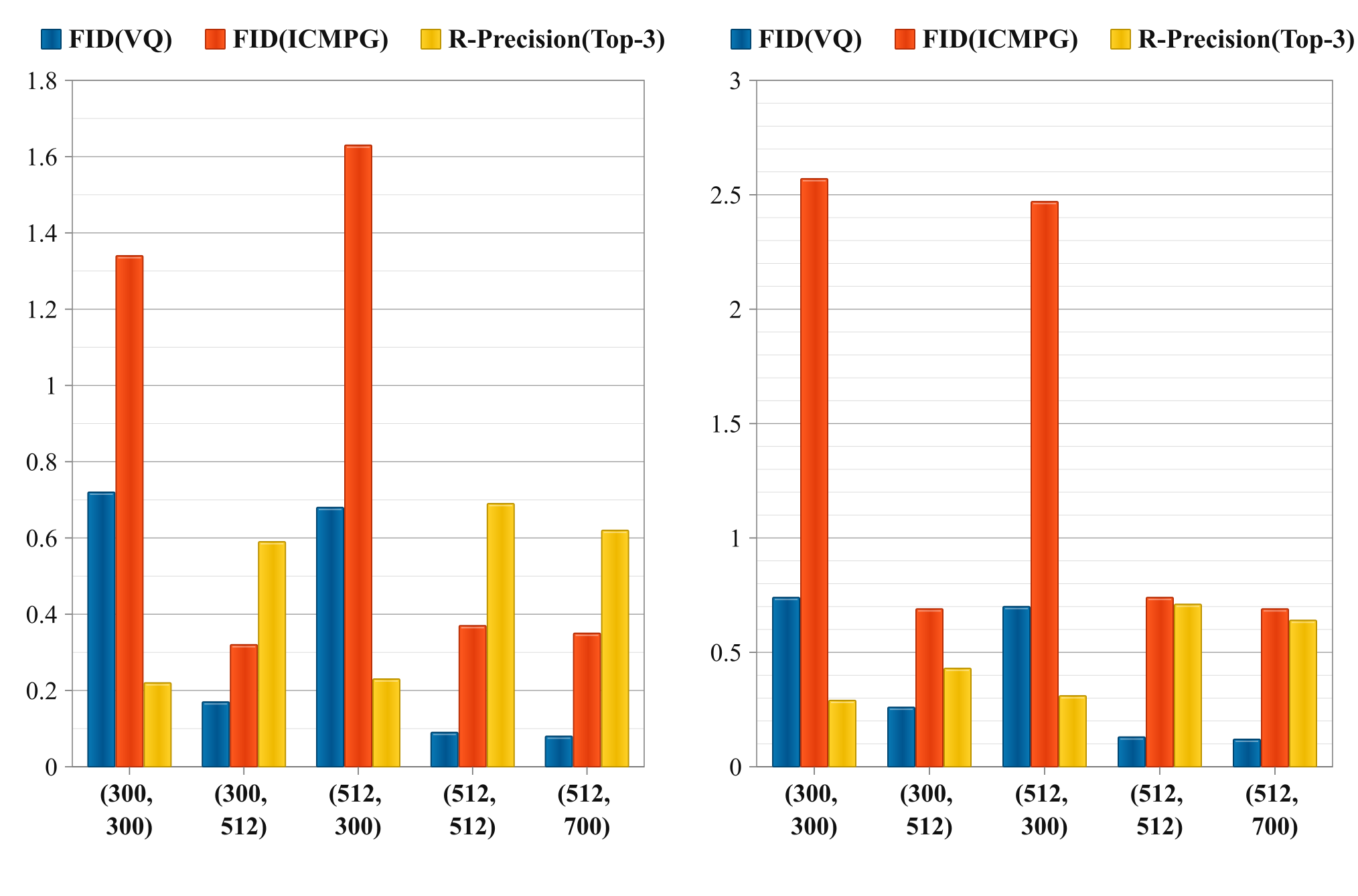}
        \caption{Effect of Codebook Size. We evaluate various codebook sizes and find that $512 \times 512$ achieves the best semantic consistency.}
        \label{fig:code}
    \end{figure}
}

\newcommand{\figablasem}{
    \begin{figure}[!t]
        \centering
        \subfloat[Iteration 0]{
            \includegraphics[width=0.36\textwidth, height=2cm]{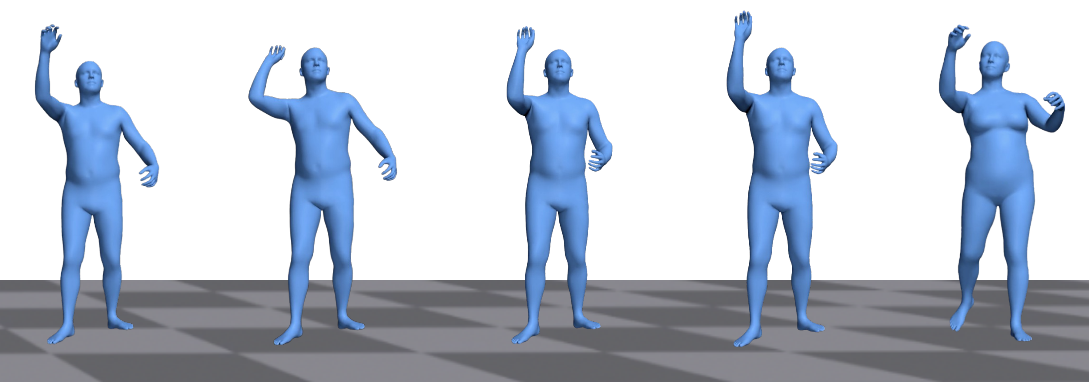}
            \label{fig:ablasem1}
        }
        \vspace{0.5cm}
        \subfloat[Iteration 5]{
            \includegraphics[width=0.36\textwidth, height=2cm]{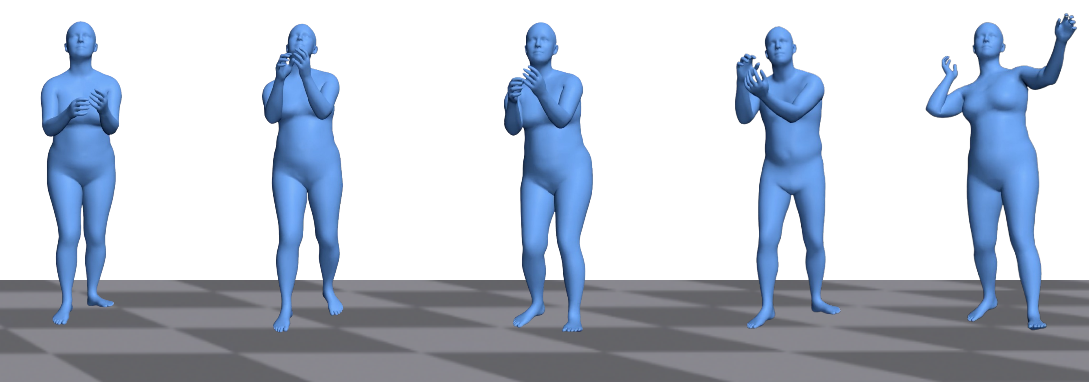}
            \label{fig:ablasem2}
        }
        \vspace{0.5cm}
        \subfloat[Iteration 10]{
            \includegraphics[width=0.36\textwidth, height=2cm]{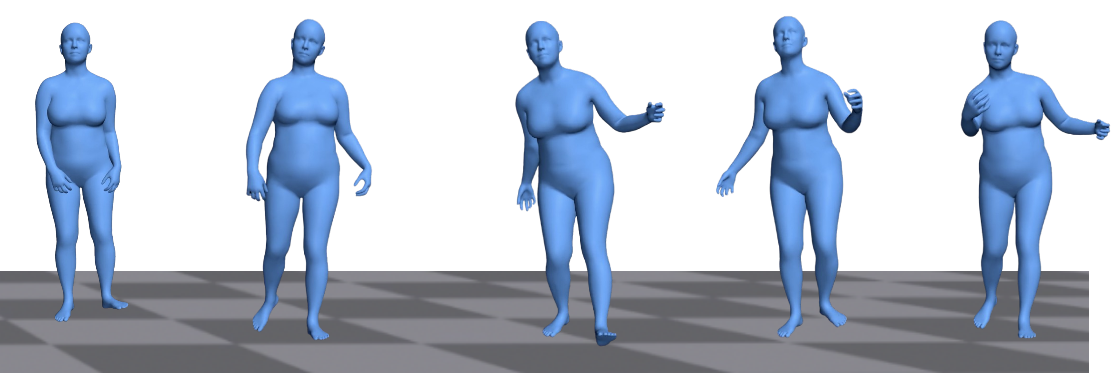}
            \label{fig:ablasem3}
        }
        \caption{Comparison of Intermediate Motions Across MPG Iterations. We compare motions during the MPG process. The results show increasing alignment between generated motions and input texts as iterations progress.}
        \label{fig:ablasem}
    \end{figure}
}

\section{Introduction}
\label{sec:intro}
\IEEEPARstart{S}{ynthesizing} human motion is increasingly important for enhancing user engagement in immersive digital environments, such as gaming, virtual reality, and cinematic production. Although existing text-to-motion methods~\cite{tevet2022human,guo2022generating,zhang2023generating,zhang2022motiondiffuse,jiang_motiongpt:_2023,zhong2023attt2m,Wang2025InstructAvatar,Bian2025MotionCraft,Peng2025HOI} have demonstrated considerable progress, their effectiveness is limited by their substantial dependence on large-scale human motion capture datasets. This reliance inherently limits their ability to generalize to unseen textual descriptions. To address this limitation, open-vocabulary motion generation tasks and related techniques have been developed. These methods acquire discriminative representations without requiring extensive data corpora, thereby enabling the synthesis of human motion from a broader range of textual inputs. Consequently, this direction has become an important research topic.

Early studies on open-vocabulary motion synthesis~\cite{tevet2022motionclip,hong2022avatarclip} trained motion sequence generators using contrastive multimodal constraints. These methods render generated motions as a series of images and align the resulting features with their textual descriptions using CLIP~\cite{radford2021learning}. However, this rendering-based paradigm is limited because projecting complex 3D kinematic data onto a 2D image plane inevitably discards important spatial details and temporal dynamics. Furthermore, the generated motions may deviate from the intended actions because of CLIP's limited ability to resolve ambiguous semantics. In this context, large language models (LLMs)~\cite{thoppilan2022lamda,achiam2023gpt,wei2022chain} have emerged as a promising methodological foundation, as subsequent studies have shown that the autoregressive capabilities of pre-trained LLMs are not confined to the language domain. This insight has facilitated the development of LLM-based human motion synthesis~\cite{sun2024prompt,liu2023plan}, which leverages the contextual understanding of LLMs to interpret open-vocabulary textual cues as sequences of detailed instructions, providing a semantically grounded framework for guiding text-to-motion generation. However, the high-level semantic plans generated by LLMs are not inherently grounded in physical reality. Although the motions are generated to semantically follow the textual instructions, they frequently violate fundamental physical laws. The resulting sequences may therefore exhibit physically implausible dynamics and fail to interact realistically with the environment.

To reproduce natural motions from text, physics-aware approaches provide physical constraints that textual supervision alone cannot offer. More specifically, PhysDiff~\cite{yuan2023physdiff} incorporates physical simulation constraints into a diffusion model and performs physical simulation on intermediate results at each diffusion step to generate motions that adapt to physical environments. AnySkill~\cite{cui2024anyskill} builds on physics-based motion imitation by integrating a semantic reward associated with open-vocabulary text into its reinforcement learning~\cite{watkins1992q,sutton1988learning} process, enabling the agent to adapt to open-vocabulary settings. However, a critical limitation of these physics-aware models is their semantic brittleness when confronted with open-vocabulary text. Because these approaches typically employ reinforcement learning to maximize generation quality in physical environments, they may insufficiently capture fine-grained instructions. As a result, such generators are less effective at synthesizing human motions for complex and imaginative prompts, for which semantic understanding is essential and physical accuracy alone is insufficient.

To reconcile high-level linguistic semantics with low-level physical compatibility in generated motions, we propose an In-Context Model Predictive Generation (ICMPG) framework inspired by Model Predictive Control (MPC)~\cite{richalet1978model,kouvaritakis2016model} for open-vocabulary human motion generation. Our method draws a conceptual analogy to the MPC framework, as illustrated in Fig.~\ref{fig:intro}. In a traditional MPC setup, the plant model represents the system dynamics and describes how the system state evolves in response to control inputs. The MPC controller uses this model to predict future states over a finite horizon, optimize a control sequence, apply the first control action to the plant, and repeat the procedure after observing the updated state. By treating the generated motion prefix as the current state and candidate future motion segments as predictions over a finite horizon, ICMPG reformulates motion synthesis as a receding-horizon optimization problem. Specifically, at each iteration, the model generates a set of candidate motion sequences. Through prediction, simulation, and comparison, it repeatedly computes physical and semantic rewards for these sequences, encouraging the generated motion to adapt to the physics engine while remaining aligned with the input semantics. The framework then selects the sequence with the highest reward. These simulation-based rewards guide the model in making subsequent motion predictions. Unlike prior reinforcement learning (RL) approaches that require task-specific policies to be trained from scratch, ICMPG performs closed-loop, inference-time optimization conditioned on demonstration examples. By combining the temporal planning capabilities of frozen LLMs with physical control, ICMPG provides more adaptable and robust open-vocabulary motion generation, addressing a key limitation of previous approaches.

To implement this methodology, we formulate the ICMPG framework with two modules: Context-Aware Motion Generation (CAMG) and Model Predictive Generation (MPG). CAMG is an LLM-based module designed to generate initial open-vocabulary motions. It uses a VQ-VAE~\cite{van2017neural} to encode human motions into discrete motion tokens and employs an external LLM, which can also be fine-tuned, to learn motion-token representations through in-context learning. CAMG enables the LLM to infer relationships between motion tokens and natural language tokens, leveraging its semantic capabilities to process open-vocabulary text. Based on these generated motions, the MPG module guides CAMG by placing the motions in a simulated physical environment and using an external semantic evaluator to compute semantic rewards for the current motion sequences, thereby informing the next CAMG generation step. MPG reformulates motion generation as a stepwise refinement process, addressing the limited controllability of single-pass outputs produced by CAMG and other open-vocabulary methods.

Our contributions are outlined as follows:
\figintro

\begin{enumerate}
    \item We introduce the In-Context Model Predictive Generation (ICMPG) framework. Unlike prior RL-based methods that rely on task-specific policy learning, ICMPG establishes a closed-loop, inference-time optimization paradigm. This approach bridges high-level semantic understanding from frozen LLMs with low-level physical realism without requiring offline policy retraining.

    \item We propose an architecture in which the LLM acts as a planner to decompose open-vocabulary instructions into atomic motion primitives. A Model Predictive Generation (MPG) module then performs online optimization over these candidates and selects the optimal candidate based on physical and semantic rewards, thereby avoiding the need for extensive offline policy training.

    \item Comprehensive evaluations on the BABEL dataset using a rigorous zero-shot protocol demonstrate that our method significantly outperforms state-of-the-art alternatives in open-vocabulary generalization. Additionally, our approach provides the flexibility to switch between LLMs or adjust training configurations for improved adaptability.
\end{enumerate}

\section{Related Work}

\subsection{Open-Vocabulary Motion Synthesis}
While a significant portion of text-to-motion research~\cite{Zeng2025Light,jiang_motiongpt:_2023,shafir2023human,zhang2023generating,zhong2023attt2m,zhang2022motiondiffuse,zhou2024emdm,chen2023executing,Wang2025InstructAvatar} focuses on generating human motions from labeled datasets~\cite{plappert2016kit,tang2023flag3d,mahmood2019amass,lin2023motion,Zhang2025MotionX}, these methods often encounter substantial challenges when processing open-vocabulary textual descriptions that differ significantly from their training distributions. To address this limitation, recent approaches~\cite{lin2023being,sun2024prompt,kalakonda2023action,yao2024moconvq} leverage LLMs to translate open-vocabulary prompts into motion representations or employ multimodal models such as CLIP~\cite{radford2021learning} and ViT~\cite{dosovitskiy2020image} for cross-modal alignment~\cite{hong2022avatarclip,tevet2022motionclip}.

However, most existing LLM-based approaches, such as \textit{Prompt, Plan, Perform}~\cite{sun2024prompt}, typically rely on open-loop planning, which separates semantic reasoning from physical execution. This separation can lead to physically inconsistent results when the generated plan exceeds the simulator's dynamic capabilities. Unlike these methods, our work introduces an in-context learning approach based on a closed-loop paradigm. By directly utilizing a frozen LLM, we iteratively refine the generated motions using feedback from physical simulation, thereby reconciling high-level semantic understanding with low-level physically constrained realism.

\subsection{Physics-Based Human Motion Control}
Simulating human motion requires strict adherence to physical laws such as gravity, friction, and contact mechanics. Various techniques have been developed to enforce these constraints through physics-aware generation and control methods~\cite{wu2025uniphys,Reda2025Physics,Rocca2025Policy,Kim2025PhysicsFC,tseng2023edge,jiang2023full,juravsky2022padl,truong2024pdp,tevet2024closd} or reinforcement learning (RL) strategies~\cite{ho2016generative,merel2020catch}. Notably, methods such as PhysDiff~\cite{yuan2023physdiff} and AnySkill~\cite{cui2024anyskill} have developed effective frameworks for generating physically plausible motions.

Nevertheless, RL-based methods often require training task-specific policies from scratch, limiting their zero-shot adaptability to open-vocabulary prompts without extensive retraining. Similarly, prior control-based methods such as PADL~\cite{juravsky2022padl} rely on offline skill libraries, thereby limiting motion diversity. In contrast, our approach draws on the principles of Model Predictive Control (MPC)~\cite{kouvaritakis2016model} by combining a general-purpose LLM with a universal, pre-trained policy. We reformulate motion generation as a receding-horizon optimization problem, shifting from ``offline policy learning'' to ``online iterative refinement.'' This formulation allows our framework to dynamically decompose complex, unseen instructions into executable primitives without task-specific tuning.

\section{Proposed Approach}

In this section, we elaborate on the In-Context Model Predictive Generation (ICMPG) framework. This approach combines the reasoning capabilities of large language models (LLMs) with the physical consistency provided by Model Predictive Control (MPC), offering a unified strategy for open-vocabulary human motion synthesis (Fig.~\ref{fig:over}). ICMPG primarily consists of two modules. First, the Context-Aware Motion Generator (CAMG) module enables the base LLM to learn motion-token representations through in-context learning and produces a set of motion candidates, denoted by $S_{m}$. Subsequently, ICMPG computes the semantic reward $\mathcal{R}_{sem}$ to evaluate the semantic similarity between each candidate motion and the input text prompt $t$, as well as the physical reward $\mathcal{R}_{phy}$ to evaluate its feasibility in a simulated physical environment. The Model Predictive Generation (MPG) module then employs these rewards to perform online refinement of the motion sequence and determine the token sequence $o_{i}$ selected at the current iteration.

\figmethod

\subsection{Context-Aware Motion Generation}
The Context-Aware Motion Generation (CAMG) module employs an LLM through in-context learning to improve its understanding of open-vocabulary text, thereby generating human motions with greater semantic consistency. CAMG first utilizes a VQ-VAE~\cite{van2017neural} to represent human motions from the datasets as discrete motion tokens. These tokens represent atomic motion primitives (e.g., 8-frame clips), enabling the LLM to compose complex actions through temporal combinations. This representation captures the semantic information of the motions in a form that can be recognized and learned by the LLM.

\subsubsection{LLM Planner}
Given the VQ-VAE output, ICMPG employs an external LLM with the following prompt to learn the corresponding motion-token representations:

\begin{tcolorbox}[colback=white, colframe=black, title=Prompts for LLM Planner]
\begin{small}
You are a text interpretation agent tasked with rephrasing the given sentence into multiple human-motion-focused descriptions.

Given a sentence, interpret it as a sequence of explicit human motions, with segments separated by commas. Each segment should start with the subject "he". Your motion space is \texttt{\textless{}atomic actions set\textgreater{}}, and you should select verbs only from this set.

Your response must be in JSON format as follows:

\{ \hspace{2em}"CoT": "Your step-by-step reasoning process.",

  \hspace{2em}"Motions": "The output sentence(s), comma-separated, each starting with 'he' and describing a human action."\}

Examples:\{...\}
\end{small}
\end{tcolorbox}

The LLM Planner constitutes the first LLM invocation in the CAMG process. It employs a pre-trained LLM that is not updated during training and does not directly generate human motions. Its role is to convert open-vocabulary inputs into structured descriptions that are easier to interpret, which we denote as the target prompt $t_p$. This process facilitates the generation of semantically consistent human motions by performing temporal decomposition and explicitly breaking down complex instructions into time-ordered atomic descriptions. It decouples high-level reasoning from low-level token generation, making open-vocabulary text inputs easier for the LLM Generator to interpret while reducing errors and noise in the computation of $\mathcal{R}_{sem}$. To develop a basic understanding of motion tokens, the LLM must capture the correspondence between simple motion tokens and natural language descriptions. For this purpose, we select examples from the CMU subset of HumanML3D and use the VQ-VAE to obtain a set of atomic action demonstrations, denoted by $A_{dem}$, which provide the LLM with initial examples of atomic actions.

\subsubsection{SemAligner}
Due to the limited context window of the LLM, ICMPG relies on relevant in-context examples provided during each inference. Therefore, we design SemAligner for semantic matching using several Transformer layers trained on the labeled motion dataset. Given an input description, SemAligner identifies semantically relevant examples from the available data and provides them as contextual demonstrations for the subsequent generation process. This selection allows the limited context window to be used more effectively while reducing the influence of irrelevant examples. SemAligner is trained with the following contrastive loss:
\begin{equation}
    L = \frac{1}{2N} \sum_{i=1}^{N} \left( y_i D_i^2 + \alpha(1 - y_i) \max(0, \mu - D_i)^2 \right)
\end{equation}
Here, $N$ denotes the batch size, and $D_i$ represents the semantic distance between the feature embeddings of the $i$-th pair. The binary indicator $y_i$ equals 1 for matched pairs and 0 otherwise. Additionally, $\mu$ denotes the margin threshold, which is set to 1 by default, while $\alpha$ controls the balance between the two loss components.

SemAligner retrieves a set of semantic demonstrations to form $S_{dem}$, which is used as a set of in-context examples for the LLM Generator. It encodes all textual data in the dataset into semantic vectors and compares them with $t_p$ to retrieve the most relevant data pairs. Compared with the atomic-action examples, these demonstrations are more complex and exhibit closer semantic correspondence to the textual descriptions, thereby narrowing the LLM's learning space and mitigating the instability of in-context learning.

\subsubsection{LLM Generator}
Provided with the SemAligner output, the LLM Generator learns the representation of motion tokens through prompting to generate a candidate set of human motions \(S_m\) based on open-vocabulary text inputs. The generator shares the model parameters with the LLM Planner while being guided by the prompts as follows:
\begin{tcolorbox}[colback=white, colframe=black, title=Prompts for LLM Generator]
\begin{small}You are to play the role of a motion tokenizer that maps text descriptions of human motions into a sequence of atomic motion tokens.

Given a sentence describing a sequence of human motions, you should identify the underlying atomic actions and output them in order as a list of motion tokens from the predefined action space: \texttt{\textless{}Atomic motion tokens list\textgreater{}}

Please refer to the following examples to generate motion tokens: \texttt{\textless{}Demonstrations\textgreater{}}. Please generate the subsequent motion tokens based on the already determined motion token sequence \texttt{\textless{}motion tokens sequence $o$\textgreater{}}.

Your response must be a valid JSON file in the following format:

\{
  "CoT": "Your step-by-step reasoning process.",

  "Candidate motion tokens": ["mt1", "mt2", ...], ["mt5", "mt8", ...]
\}

Examples:\{...\}\end{small}

\end{tcolorbox}
It emphasizes fully leveraging the semantic understanding capabilities of the LLM and generates human motions entirely through an in-context learning approach.

The LLM Generator receives prompt information provided by the VQ-VAE, LLM Planner, and SemAligner and performs in-context learning to generate candidate motion-token sequences that can be decoded by the VQ-VAE decoder. The resulting set of motion candidates is further evaluated during the Model Predictive Generation (MPG) process to select the optimal motion-token sequence $o_i$, which is then returned to the LLM Generator for subsequent refinement. The role of the LLM Generator in the CAMG and MPG processes is further described in Sections~\ref{s1} and~\ref{s2}.

\subsection{Model Predictive Generation}
Despite its ability to align open-vocabulary concepts, the CAMG workflow relies on the effectiveness of in-context learning, and its generated motions may be less stable than those produced by supervised methods. Therefore, we design the Model Predictive Generation (MPG) process to further refine the generated human motions. Specifically, MPG draws on the principles of Model Predictive Control (MPC) by treating motion generation in CAMG as a model prediction process. It evaluates the corresponding physical and semantic rewards using the physical environment and semantic encoder, respectively, to guide motion-sequence generation.

\subsubsection{Control Sequence Design}
In the original CAMG process, the LLM Generator receives outputs from the SemAligner and LLM Planner, together with data extracted from motion datasets, as prompts. To achieve stepwise motion refinement, we incorporate the generated motion-token sequence \(o\) into the prompts. Specifically, \(o\) is initially empty. The motion-token sequences in the candidate set $S_m^{(i)}$, generated at the \(i\)-th iteration, are evaluated in the physical environment to calculate the physical reward \(\mathcal{R}_{phy}\) and by the semantic encoder to compute the semantic reward \(\mathcal{R}_{sem}\). We select the short-term motion sequence \(o_i\) with the highest reward from $S_m^{(i)}$ and append it to \(o\). The updated sequence is then included in the prompts to guide generation at the \((i+1)\)-th iteration. Note that \(o_i\) is a sequence rather than a single token, allowing its length to be adjusted flexibly to provide an appropriate prompt configuration for the LLM Generator.

\subsubsection{Reward Function Design}
\label{s1}
As shown in Fig.~\ref{fig:over}, we first employ the pre-trained policy $\pi$ provided by PHC~\cite{luo2023perpetual} to simulate motions based on existing motion datasets. Our approach focuses on generating human motions that are compatible with physics-based environments rather than training a new policy to make unrealistic or poorly structured motions physically feasible. Leveraging this external policy encourages the generated motions to conform to physical constraints, allowing the framework to focus on high-level semantic and temporal coherence. For each motion sequence \(m\) in \(S_m^{(i)}\), MPG evaluates it using the policy \(\pi\) in the Isaac Gym~\cite{makoviychuk2021isaac} environment and computes the physical reward \(\mathcal{R}_{phy}\) over a prediction horizon of \(H\) motion tokens as follows:
\begin{equation}
\mathcal{R}_{phy}(m)
=
\frac{
\sum_{j=1}^{H}\gamma^{j-1}R(o+m_{1:j})
}{
\sum_{j=1}^{H}\gamma^{j-1}
},
\end{equation}
\begin{equation}
S_q = \mathop{\text{Top-}K}\limits_{m_{1:H}, m \in S_m^{(i)}} \mathcal{R}_{phy}(m)
\end{equation}
where \(R \in [0,1]\) denotes the normalized reward computed using the reinforcement learning objective associated with policy \(\pi\). The set \(S_q\) contains the \(\mathrm{top}\text{-}K\) motion-token sequences in \(S_m^{(i)}\) with the highest physical rewards \(\mathcal{R}_{phy}\). The plus sign denotes motion concatenation, whereby a subsequent segment is appended to the preceding sequence. The notation \(m_{1:j}\) represents the subsequence formed by the first \(j\) tokens of \(m\). Here, \(\gamma\) is the discount factor, and \(\gamma^{j-1}\) assigns position-dependent weights to the rewards over the prediction horizon.

Beyond the physical signals, MPG also renders the motion sequence $m \in S_m^{(i)}$ into images using the SMPL~\cite{loper2023smpl} method, and then selects some key frames to form the set \(G\), which is fed into the external image encoder \(C_{im}\) to compute their semantic vector. Rendering the motion into textureless SMPL meshes serves as a canonicalization step. By removing background distractions inherent in raw video data, we reduce appearance variation and encourage the encoder to focus on pose-related cues. This allows the encoder to focus exclusively on motion dynamics, effectively mitigating the domain gap between real videos and synthesized avatars. Afterwards, the text encoder \(C_{text}\) is used to calculate the semantic vector of the output \(t_{p}\) from the LLM Planner, and the cosine similarity between the two is computed as the reward:
\begin{equation}
\mathcal{R}_{sem}(m) =
\frac{
C_{text}(t_{p}) \cdot \sum_{j=1}^{N_G} C_{im}(G_{j})
}{
\|C_{text}(t_{p})\| \left\|\sum_{j=1}^{N_G} C_{im}(G_{j})\right\|
}
\end{equation}
where $N_G$ represents the number of selected key frames in $G$, which is set to 4 by default and $\|\cdot\|$ denotes the L2 norm.

\subsubsection{Online Motion Refinement}
Given the physical and semantic reward signals, the MPG module combines them to infer the total reward information. We select the candidate \(q \in S_q\) with the highest total reward and use its horizon segment \(q_{1:H}\) as \(o_i\). The specifics are as follows:

\begin{equation}
\mathcal{R}_{total} = \mathcal{R}_{phy} + \beta\mathcal{R}_{sem}
\end{equation}
\begin{equation}
o_i = \left(\mathop{\arg\max}\limits_{q, q \in S_q} \mathcal{R}_{total}(q)\right)_{1:H}
\end{equation}
where $\beta=1$ is a hyperparameter balancing the physical and semantic rewards. To mitigate potential hallucinations, we further implement a re-generation protocol: if the semantic reward $\mathcal{R}_{sem}$ of the selected $o_i$ falls below a threshold $\epsilon$, the current generation step is re-triggered, where $\epsilon$ is selected on the validation set. The execution of online motion refinement draws inspiration from the iterative cycle of the MPC controller. We reframe the task of motion synthesis as receding-horizon optimization with the goal of progressively building a physically and semantically coherent motion sequence. The process mirrors the loop of an MPC controller at each generation step.
\begin{enumerate}

    \item State Measurement: The ``current state'' of our system is the motion sequence, $o$, that has been generated so far. Initially, this sequence is empty.

    \item Prediction and Optimization: The LLM Generator acts as our predictive model. It generates a set $S_m^{(i)}$ of candidate motion sequences by extending from the current motion prefix $o$. These candidates represent possible futures over a prediction horizon. We then evaluate each candidate using physical and semantic rewards to identify the optimal segment.

    \item Control Action: Our ``control action'' is to select the single best motion in $S_m^{(i)}$ based on its physical and semantic rewards and append it to our current sequence $o$.

    \item Repeat: With the updated sequence $o$ as the new state, the entire cycle repeats, progressively extending the motion. Once the generation is complete, the final sequence $o$ is decoded into our human motion results.

\end{enumerate}

While the underlying principle is analogous, our online refinement adapts the MPC concept to generative AI. A traditional MPC controller typically optimizes a sequence of low-level control actions, such as torques or voltages, to determine an optimal trajectory. In contrast, our method operates at a higher level of abstraction. Using the LLM, we generate high-level candidate motion sequences over the upcoming horizon and evaluate them to select the optimal one.

\subsubsection{Fine-Tuning MPG Module}
\label{s2}
To further improve the performance of ICMPG, we explore a variant based on Low-Rank Adaptation (LoRA) fine-tuning~\cite{hu2022lora}. In this setting, we reformulate the CAMG generation process to produce a single motion-token sequence at each iteration rather than a set of motion candidates. We also train a differentiable world model, implemented as a neural network that approximates the PHC simulator. This model allows gradients from the synthesized-motion reconstruction loss to backpropagate directly to the upstream components. The world model is trained iteratively by minimizing the discrepancy between its predictions and actual simulation trajectories sampled from a replay buffer. These modifications reduce the computational overhead during training, improving the efficiency of the fine-tuning process.

Within this differentiable framework, we formalize the generation and optimization pipeline at iteration $i$ using a fixed horizon $H$. The CAMG module conditions on the processed text prompt $t_p$ produced by the LLM Planner and the preceding motion history $o$, which is initially empty, to synthesize a single motion-token sequence $m$. Subsequently, the optimization objective incorporates both physical simulation constraints and semantic alignment to minimize the discrepancy between $m$ and the ground truth. The loss function is defined as follows:
\begin{equation}
\mathcal{L}_{phy} = -\log\left(\sum_{j=1}^{H}\gamma^{j-1}R(o + m_{1:j})\right)
\end{equation}
\begin{equation}
\mathcal{L}_{sem} = 1 - \mathcal{R}_{sem}(m)
\end{equation}
\begin{equation}
\mathcal{L}_{total} = \lambda_{1} \mathcal{L}_{MSE}(o + m_{1:H}, M) + \lambda_{2}\mathcal{L}_{phy} + \mathcal{L}_{sem}
\end{equation}
where $M$ denotes the set of motion-token sequences corresponding to the textual inputs in the dataset, $\mathcal{L}_{MSE}$ represents the mean squared error loss, and $\lambda_1$ and $\lambda_2$ are weighting coefficients that are set to 1 by default. The LoRA-fine-tuned ICMPG framework follows a conventional training strategy and generates a single motion-token sequence per iteration. The primary role of LoRA fine-tuning is to provide the LLM Generator with a basic representation of motion tokens, thereby enabling more effective motion generation in the early stages than methods that rely solely on prompting.

\section{Experiments}
In this section, we present an evaluation of the ICMPG framework on both standard and open-vocabulary text-to-motion generation tasks. We conduct experiments on three datasets, HumanML3D, KIT-ML, and BABEL, to assess the model's performance from perspectives including motion quality, semantic consistency, and physically constrained realism. We further conduct ablation studies to analyze key components such as the VQ-VAE codebook, LLM backbone configuration, and reward function design.

\subsection{Datasets and Experimental Settings}
We conduct evaluations on three benchmark datasets: HumanML3D, KIT-ML, and BABEL. HumanML3D extends the AMASS dataset~\cite{mahmood2019amass} by providing detailed textual descriptions for each motion sequence. It contains 14,616 motion clips paired with 44,970 sequence-level annotations describing the corresponding motions in natural language. KIT-ML provides 6,353 textual descriptions aligned with 3,911 motion sequences, offering a compact yet semantically rich benchmark for text-to-motion generation. For Tables~\ref{tab:com2} and~\ref{tab:com1}, we follow the official dataset-specific training and evaluation protocol of each method. Unless otherwise stated, the BABEL evaluation uses models trained on HumanML3D without BABEL-specific fine-tuning. For HumanML3D, we follow the standard partition, allocating 80\% of the data for training, 5\% for validation, and 15\% for testing. For BABEL and KIT-ML, the datasets are used in their entirety for testing unless otherwise specified.

While HumanML3D and KIT-ML are widely used to evaluate standard text-to-motion generation models, their limited lexical diversity does not fully reflect the challenges of open-vocabulary settings. In contrast, BABEL is a large-scale dataset containing natural language annotations of motions in motion capture (mocap) sequences. It provides semantic labels for approximately 43 hours of mocap data derived from AMASS at two levels of abstraction. Sequence-level annotations describe the overall content of each clip, whereas frame-level annotations provide fine-grained descriptions of motions over specific temporal intervals. Each frame-level annotation is temporally aligned with its corresponding motion segment, allowing multiple motions to be annotated simultaneously. Our statistical analysis shows that approximately \textbf{24\%} of action verbs and \textbf{39\%} of object nouns in the test set are unseen during training, providing a challenging open-vocabulary setting.

\tabnormall
\tabnormal

\subsection{Implementation Details}
In the VQ-VAE module, input motion sequences are encoded into atomic codebook indices using 8-frame segments, whereas decoding is performed globally. We employ a non-autoregressive Transformer decoder with bidirectional attention to process the entire token sequence simultaneously. This global decoding strategy promotes smooth transitions between atomic segments and reduces boundary artifacts commonly observed in clip-wise decoding methods. For semantic alignment, SemAligner uses a BERT-Large-Uncased backbone fine-tuned through contrastive learning on HumanML3D. We set the weighting coefficient $\alpha$ to 0.2 to balance generalization and robustness to noise. The trainable modules are optimized using AdamW~\cite{loshchilov2017decoupled} with momentum parameters $(\beta_1^{\mathrm{Adam}}, \beta_2^{\mathrm{Adam}}) = (0.9, 0.99)$. For the LLM backbone, we employ \textit{LLaMA-3.1-70B-Instruct}~\cite{grattafiori2024llama} for both the Planner and the non-fine-tuned Generator to leverage its reasoning capability. For the LoRA~\cite{hu2022lora} fine-tuning stage, we instead adopt \textit{LLaMA-3.1-8B-Instruct} to reduce computational overhead while maintaining generation quality. To construct the prompts, we extract explicit atomic action labels (e.g., ``walking'' and ``jumping'') from the CMU~\cite{hodgins2015cmu} subset of HumanML3D. These demonstrations, denoted by $A_{dem}$, provide prior examples for mapping semantic concepts to specific motion-token sequences. For reward inference, the physical policy is initialized using PHC~\cite{luo2023perpetual}, which reports a 100\% simulation success rate on AMASS within Isaac Gym~\cite{makoviychuk2021isaac}. Meanwhile, we employ ActionCLIP~\cite{wang2021actionclip} as the semantic evaluator to quantify the alignment between synthesized motions and the input text.

\figopenvocab
\figcontrast

In the MPG process, we set the prediction horizon to $H=4$ motion tokens, generate 10 candidates at each step, and retain the $\mathrm{top}\text{-}K$ sequences for the subsequent iteration. These retained sequences provide candidate continuations for the next refinement step. The discount factor in Eq.~(1) is set to 0.95. For probabilistic generation, we use a temperature of 0.6 and a $\mathrm{top}\text{-}p$ sampling threshold of 0.9 to encourage diversity among the generated candidates. The prompt incorporates in-context demonstrations retrieved by SemAligner, and the physical simulation operates at 30~Hz. The ICL-based ICMPG requires more than 24~GB of VRAM for inference, whereas the LoRA fine-tuning variant, which uses an 8B model, requires at least 40~GB of VRAM during fine-tuning. Further details on computational efficiency are provided in Appendix~D.

\subsection{Evaluation Metrics}
Our experiments evaluate the generated motions from three perspectives: motion generation capability, semantic consistency, and physically constrained realism. These complementary criteria assess the overall quality of the generated motions at both the distributional and individual-sequence levels.

\paragraph{Generation Capability}
To assess motion generation quality, we employ the Fréchet Inception Distance (\textit{FID}), which quantifies the distributional discrepancy between generated motions and ground-truth sequences. To measure motion diversity, we randomly sample multiple pairs of generated motions and compute their $\ell_2$ distances as the diversity metric. This metric reflects the variation among the generated samples. Together, these metrics evaluate the model's ability to generate realistic and diverse human motion sequences.

\paragraph{Semantic Consistency}
For semantic evaluation, we adopt \textit{R-Precision}, a widely used retrieval-based metric for measuring motion-text correspondence. Specifically, we combine the original input text with 32 randomly selected textual descriptions and determine whether the original description appears among the $\mathrm{top}\text{-}n$ texts most relevant to the generated motion. We report Top-1 and Top-3 results. The evaluation is conducted over multiple motion-text pairs to improve the reliability of the measured performance. For open-vocabulary evaluation, we also report the \textit{CLIP\_Score} to provide an additional measure of text-motion semantic alignment.

\tabopen
\figablasem
\figlong
\figcode

\paragraph{Physically Constrained Realism}
To evaluate the physical realism of the generated motions, we follow the protocol proposed in PhysDiff~\cite{yuan2023physdiff} and compute several physics-based error metrics in a simulated environment. Specifically, \textit{Penetrate} measures ground penetration depth, \textit{Float} evaluates unnatural floating behavior, and \textit{Skate} quantifies foot sliding artifacts. These individual metrics are aggregated into a composite score, denoted as \textit{Phys-Err}, expressed in millimeters. The simulation environment serves as a realistic test bed for assessing motion feasibility under physical constraints.

\subsection{Comparison with State-of-the-Art Methods}
We compare ICMPG with existing text-to-motion generation methods on three benchmark datasets: HumanML3D, KIT-ML, and BABEL. We divide the evaluation into two categories: standard text-to-motion generation and open-vocabulary settings. The first category focuses on standard textual descriptions and assesses the model's ability to generate realistic and coherent motion sequences. The second category evaluates its generalization to unseen or rare textual descriptions, reflecting its robustness in more complex scenarios.

\subsubsection{Evaluation on Standard Textual Descriptions}
We compare ICMPG with conventional text-to-motion methods to assess whether the in-context-learning-based framework can achieve performance comparable to leading approaches. As detailed in Tables~\ref{tab:com2} and~\ref{tab:com1}, we evaluate ICMPG against several representative methods under the corresponding benchmark protocols. A key component of our framework is the VQ-VAE module in the CAMG pipeline. Experimental results show that our VQ-VAE achieves a low \textit{FID} and competitive retrieval and physical metrics, indicating strong reconstruction quality and a close approximation to the distribution of real motion data. These results demonstrate its ability to reconstruct and preserve essential motion features.
\tabablaLLM

We further analyze the overall performance of ICMPG in terms of motion quality and semantic alignment. Under the baseline configuration using only in-context learning, the Top-1 and Top-3 \textit{R-Precision} scores of ICMPG are only 0.026 and 0.025 below those of the best-performing method on the HumanML3D test set, respectively. The corresponding differences on KIT-ML are 0.018 and 0.024. With LoRA fine-tuning, these performance gaps are further reduced, and ICMPG achieves competitive \textit{FID} scores, indicating strong motion generation quality. It also outperforms the non-physics-aware baselines in physically constrained realism and achieves the best \textit{Phys-Err} performance under the LoRA configuration.

\subsubsection{Evaluation on Open-Vocabulary Textual Descriptions}
We evaluate the model's ability to generalize to open-vocabulary textual descriptions while generating physically plausible motions. We conduct experiments on the BABEL dataset and compare ICMPG with open-vocabulary motion generation methods. As shown in Table~\ref{tab:com}, the baseline ICMPG outperforms AvatarCLIP and MotionCLIP in \textit{FID}, \textit{CLIP\_Score}, and \textit{R-Precision}. With LoRA fine-tuning, it further surpasses AnySkill across these metrics. In physics-based simulation, ICMPG achieves success rates and \textit{Phys-Err} scores comparable to those of AnySkill without task-specific reinforcement learning. These results demonstrate that ICMPG performs well in both semantic alignment and physical realism when processing open-vocabulary textual descriptions.

\subsection{Qualitative Comparison}
We qualitatively evaluate the proposed ICMPG framework under open-vocabulary textual descriptions, physically constrained realism, and long-sequence generation. These aspects are examined through visual comparisons and simulation-based analyses to illustrate the ability of our method to generate expressive and realistic human motions.

\subsubsection{Evaluation on Open-Vocabulary Inputs}
We visualize motions generated from open-vocabulary inputs using the SMPL model (Fig.~\ref{fig:qualitative_results}). In the first case, the motions generated by MotionCLIP and AnySkill do not capture the semantics of ``ballroom dance'' or dance movements. In contrast, ICMPG generates motions that are semantically aligned with the textual description. Moreover, ICMPG captures the implicit meaning of ``ballroom dance,'' producing sequences that reflect the semantics of dancing and exhibit cha-cha-specific characteristics. In the second example, where the target action is jumping with raised hands, ICMPG generates a motion that reflects the semantics. These results indicate that ICMPG produces semantically meaningful outputs for open-vocabulary textual descriptions.
\figablation

%ballroom dance in pattern series
%hands up high jump

%sprint backwards

\subsubsection{Evaluation on Physically Constrained Realism}
To assess the physical realism of the generated motions, we conduct visual comparisons in the Isaac Gym simulation environment. We select two representative tasks: ``perform a squat'' and ``sprint backwards.'' These tasks pose substantial challenges in terms of balance and physical constraints. As illustrated in Fig.~\ref{fig:physs}, MotionCLIP generates physically implausible motions characterized by floating behavior and poor foot-ground contact. AnySkill also struggles to maintain stability, resulting in simulation failure during squatting and limb entanglement during running transitions. In contrast, ICMPG iteratively refines the generated motions by adjusting their depth and posture to improve physically constrained realism. These results indicate that ICMPG not only aligns with the input texts but also generates motions suitable for physical simulation.

\subsubsection{Evaluation on Long-Sequence Generation}
We further evaluate ICMPG using long-form textual descriptions containing multiple complex semantic elements. Through its in-context learning strategy, our framework decomposes such descriptions into motion-focused prompts, enabling the iterative generation of extended and coherent motion sequences.

A key challenge is generating high-quality transitions between motion segments derived from different semantic components. As shown in Fig.~\ref{figlong}, ICMPG generates coherent motions in the first two examples. More notably, it produces smooth and stable motion sequences even when the underlying semantics undergo substantial transitions. This capability is supported by the MPG process, which reformulates motion generation as a temporal refinement procedure. Specifically, ICMPG iteratively infers the next motion segment based on previously generated content while maintaining physically constrained realism within the simulation environment. This mechanism enables our method to perform effectively in long text-to-motion generation tasks, particularly those requiring long-term semantic consistency and physical realism.

\begin{table*}[h]
    \caption{Comparison on challenging benchmarks. \textbf{UniPhys} excels in physics but is not designed for text-level semantic generation; \textbf{MotionCraft} is semantically strong but physically unstable. \textbf{ICMPG} Achieves the Highest MM-Hit Among the Compared Methods While Substantially Reducing Mod-Cost Relative to Generative Baselines, Demonstrating a Favorable Balance Between Semantic Generalization and Physical Reliability.}
    \label{tab:sota_2025}
    \centering
    \resizebox{\linewidth}{!}{
    \begin{tabular}{l|cc|cc|c}
        \toprule
        & \multicolumn{2}{c|}{\textbf{Semantic (MotionMillion)}} & \multicolumn{2}{c|}{\textbf{Physical (PP-Motion)}} & \\
        Method & \textbf{MM-Hit@1} $\uparrow$ & Diversity $\rightarrow$ & \textbf{Mod-Cost} $\downarrow$ & Penetration $\downarrow$ & Paradigm \\
        \midrule
        MotionGPT~\cite{jiang_motiongpt:_2023} & 68.4\% & 9.52 & 10.21 & 2.14 & Generative \\
        \textbf{UniPhys (ICCV'25)}~\cite{wu2025uniphys} & 58.7\% & 8.91 & \textbf{0.82} & \textbf{0.00} & Physics-Ctrl \\
        \textbf{MotionCraft (AAAI'25)}~\cite{Bian2025MotionCraft} & 74.1\% & \textbf{9.85} & 7.92 & 1.35 & Generative \\
        \midrule
        \textbf{ICMPG (Ours)} & \textbf{76.3\%} & 9.78 & 1.85 & 0.18 & \textbf{Hybrid} \\
        \bottomrule
    \end{tabular}
    }
\end{table*}

\subsubsection{Comparative Evaluation on Challenging Benchmarks}
To evaluate the model's zero-shot generalization and physical realism, we incorporate the MotionMillion benchmark~\cite{fan2025go} to assess semantic alignment on million-scale prompts and the PP-Motion metric~\cite{zhao2025pp} to quantify physical fidelity using simulation-based costs. As shown in Table~\ref{tab:sota_2025}, UniPhys~\cite{wu2025uniphys} achieves a low Mod-Cost of 0.82 but obtains an MM-Hit of 58.7\%, limiting its performance on open-vocabulary tasks. Conversely, MotionCraft~\cite{Bian2025MotionCraft} supports diverse motion synthesis but exhibits a relatively high Mod-Cost, reflecting artifacts such as foot sliding and postural instability. ICMPG balances these objectives and achieves a higher MM-Hit than MotionCraft (\textbf{76.3\%}), while the MPG module maintains a Mod-Cost of 1.85. These results demonstrate that ICMPG provides a favorable balance between semantic fidelity and physical realism. To complement these objective metrics, we conduct a double-blind human perceptual study ($N=30$) under out-of-distribution (OOD) semantic shifts. The perceptual results are consistent with the objective findings, indicating that ICMPG mitigates contextual overload in long-horizon instructions and performs favorably against LLM-based and physics-based baselines in semantic alignment and kinematic naturalness (see the Supplementary Material for detailed statistics).

\subsection{Ablation Study}
We conduct ablation studies to evaluate the contribution of each key component in our framework, including the VQ-VAE codebook design, LLM variants, LLM Planner, and reward function design. These experiments are performed on the HumanML3D and BABEL datasets and aim to provide a clear understanding of how each component affects the overall performance of the framework.

\subsubsection{Analysis of Codebook Design}
We analyze the impact of codebook size on the performance of ICMPG because it directly influences the quality and capacity of the motion representations. The codebook length $l$ determines the dimensionality of the encoded vectors and thus affects the capacity of the latent motion space. Meanwhile, the codebook width $w$ corresponds to the total number of motion tokens and influences the LLM Generator's ability to interpret and utilize these tokens effectively.

As shown in Fig.~\ref{fig:code}, increasing $w$ beyond 512 results in only marginal improvements. For example, on the HumanML3D dataset, the difference in \textit{FID} between codebook widths of 512 and 700 is less than 0.02, whereas the difference in \textit{R-Precision} is 0.06. Human motion sequences are typically represented using 22 joints, three spatial coordinates, and a variable number of frames. In our experimental setup, motions are encoded into 8-frame segments, resulting in 528 scalar values per segment, which closely matches the commonly used codebook length of 512. When $l=512$, a codebook with width $w=700$ achieves a lower VQ-VAE \textit{FID} than one with $w=512$, indicating better reconstruction capability. However, it performs worse in terms of \textit{R-Precision}. This result may be attributed to the increased number of motion tokens associated with a larger $w$, which can introduce redundancy and hinder the LLM from capturing fine-grained semantic distinctions between motion tokens.

\subsubsection{Analysis of LLM Variants}
Our method generates human motions from open-vocabulary textual descriptions by employing large language models (LLMs) as core components. We adopt an in-context learning strategy without fine-tuning the LLM, aiming to provide a plug-and-play design that can flexibly incorporate future advances in LLM technology.

To assess the impact of different LLM choices on model performance, we implement two variants: one based on \textit{LLaMA-2-70B-Instruct}~\cite{touvron2023llama} and the other based on \textit{Phi-4-14B}~\cite{abdin2024phi}, both of which are used in their original, unmodified forms. As shown in Table~\ref{tab:LLM}, the LLaMA-2-based variant exhibits noticeable degradation in generation quality and semantic consistency. These results indicate that the choice of backbone affects the performance and effectiveness of our framework, with more capable LLMs generally providing improved semantic alignment and motion coherence. Nevertheless, we select LLaMA-3.1 as our baseline because of its superior performance in physically constrained realism, particularly in terms of \textit{Phys-Err}. Our framework is model-agnostic and is expected to benefit from future improvements in LLMs, allowing continued enhancement of open-vocabulary motion generation.

\subsubsection{Impact of LLM Planner}
The LLM Planner serves an important function in the CAMG module by performing \textit{semantic normalization} and \textit{temporal decomposition}. By transforming unstructured open-vocabulary instructions into structured atomic action descriptions, the Planner decouples high-level semantic reasoning from low-level motion-token synthesis. To evaluate this architectural design, we conduct a targeted ablation study using a curated subset of 500 complex, long-horizon prompts from the BABEL dataset. This setting evaluates the model's ability to maintain semantic fidelity under challenging conditions. Furthermore, to investigate the relationship between the modular design of the framework and the reasoning capacity of the underlying foundation model, we introduce Gemini 3 Pro~\cite{gemini3pro2026} as a comparative backbone. This comparison allows us to evaluate the ``plug-and-play'' nature of ICMPG and examine how advances in LLM capabilities affect the contribution of a dedicated planning module.

\begin{table}[h]
    \caption{Ablation Study on the LLM Planner Using a Subset of 500 Complex BABEL Prompts. The results indicate that the Planner is important for the standard ICMPG configuration, while its contribution becomes less pronounced as the semantic capability of the backbone increases. The ``Full'' version consistently outperforms the ``w/o Planner'' baseline across all evaluated backbone configurations.}
    \label{tab:planner_ablation}
    \centering
    \begin{tabular}{l|cc}
        \toprule
        Method & FID $\downarrow$ & Success Rate $\uparrow$ \\
        \midrule
        Ours (Default, w/o Planner) & 0.932 & 53.6\% \\
        Ours (Default, Full) & 0.614 & 74.9\% \\
        \midrule
        Ours (Gemini 3 Pro, w/o Planner) & 0.536 & 84.7\% \\
        Ours (Gemini 3 Pro, Full) & \textbf{0.514} & \textbf{86.3\%} \\
        \bottomrule
    \end{tabular}
\end{table}

As presented in Table~\ref{tab:planner_ablation}, removing the LLM Planner from our default backbone (LLaMA-3.1-70B) results in substantial performance degradation, with the success rate decreasing from 74.9\% to 53.6\% and the FID increasing to 0.932. These results indicate that standard open-weight models may struggle to simultaneously parse complex semantic structures and generate coherent motion tokens. Conversely, when using Gemini 3 Pro, the performance gap between the ``w/o Planner'' and ``Full'' configurations becomes substantially narrower. This observation provides two main insights. First, the Planner provides important reasoning support for current mainstream models. Second, ICMPG exhibits compatibility with stronger foundation models. As the reasoning capabilities of foundation models improve, our framework can more effectively handle the combined tasks of semantic parsing and motion generation. These findings demonstrate the flexibility of ICMPG and its potential to benefit from future advances in LLM technology. Visual comparisons that further support these findings are provided in the Supplementary Material.

\subsubsection{Analysis of Reward Function Design}
The MPG module integrates a reward mechanism that guides motion selection using both physical simulation and semantic alignment. Balancing these two aspects is important for generating high-quality and physically realistic motion sequences.

To evaluate the influence of the reward function, we systematically vary the hyperparameter $\beta$ to control the weight of the semantic reward during the refinement process. As shown in Fig.~\ref{fig:m}, ICMPG consistently outperforms the baseline without MPG across all evaluated settings. These results demonstrate the effectiveness of MPG in improving sequence coherence and contextual appropriateness. As illustrated in Fig.~\ref{fig:ablasem}, the generated motion progressively aligns with the input semantics over successive MPG iterations, qualitatively illustrating the iterative refinement process. Increasing $\beta$ leads to improvements in \textit{R-Precision}, suggesting that stronger semantic weighting improves alignment with the textual descriptions. Conversely, the \textit{Phys-Err} metric exhibits a non-monotonic trend, initially decreasing and subsequently increasing. This behavior reflects a trade-off between semantic consistency and physically constrained realism, as excessive semantic emphasis can weaken adherence to physical constraints. At lower values of $\beta$, assigning a moderate semantic weight helps the refinement process select more semantically plausible motion tokens and reduces the generation of semantically inconsistent sequences. However, as the semantic weight increases further, the physically constrained realism enforced by MPG is gradually weakened, leading to the observed increase in \textit{Phys-Err}. Overall, the reward function provides a balance between semantic fidelity and physical realism.

We further assess the impact of different semantic encoders used during semantic reward inference. Specifically, we implement and compare several variants based on CLIP-ViT-B/16, CLIP-ViT-B/32, and ActionCLIP. In addition, we retain a model that employs only the CAMG module as a baseline for comparison. The results indicate that all variants exhibit a consistent performance trend with respect to $\beta$, demonstrating the stability of the MPG mechanism across different semantic models. The variant incorporating ActionCLIP achieves the best overall performance, likely because its additional fine-tuning on human motion data improves its ability to align motion sequences with textual descriptions. Furthermore, to evaluate the robustness of the composite pipeline, we conduct targeted noise-injection experiments, as detailed in the Supplementary Material. Even under retrieval perturbations with 50\% noise or semantic reward corruption, ICMPG exhibits gradual performance degradation, maintains a Top-3 \textit{R-Precision} above 38.5\%, and avoids severe kinematic failures. These results provide additional evidence of the robustness of the closed-loop MPG filtering process.

\section{Conclusion}
In this paper, we presented ICMPG, a framework for generating physically plausible motions from open-vocabulary text by combining general-purpose large language models (LLMs) with model-predictive refinement. By reformulating motion generation as a receding-horizon optimization problem, ICMPG avoids the need for task-specific policy training. Specifically, the framework first employs LLMs to learn motion-token representations through in-context learning and then uses MPG to iteratively refine the generated sequences according to semantic and physical feedback, thereby improving semantic coherence and physical feasibility.

Experimental results demonstrate that ICMPG achieves strong semantic alignment while maintaining physically plausible motion generation, with competitive or superior performance across the evaluated standard and open-vocabulary benchmarks. These findings indicate the potential of closed-loop, inference-time refinement for integrating semantic reasoning with physical constraints. In future work, we aim to further leverage the reasoning and interaction capabilities of LLMs to advance controllable multimodal human motion generation.

\section*{Acknowledgments}
This work was supported in part by the National Natural Science Foundation of China under Grants 62322608, 62572498, and 62506180, and in part by the PCL Major Key Project under Grant PCL2025A17-2. The authors would like to thank National Supercomputer Center in Guangzhou for providing high performance computational resources.
\appendix

% \section{Supplementary Experiments and Analysis}

In this appendix, we present extended experimental analyses to validate the design choices of the ICMPG framework. We specifically focus on the robustness of our semantic alignment module, the reliability of our reward oracle under domain shift, the expressiveness of the motion codebook for out-of-distribution (OOD) data, and a transparent analysis of the computational trade-offs inherent to our inference-time optimization paradigm.

\begin{table}[h]
    \caption{Performance comparison of retrieval strategies on the \textbf{OOD BABEL dataset}. The substantial lead of SemAligner over Pure BERT confirms that domain-specific alignment is required for generalization, while the narrow gap between our model and the Oracle variant (trained on BABEL) validates the robustness of our approach.}
    \label{tab:semaligner_ablation}
    \centering
    \begin{tabular}{l|cc}
        \toprule
        Retrieval Strategy & R-Precision (Top-3) $\uparrow$ & FID $\downarrow$ \\
        \midrule
        Random Selection & 0.271 & 0.876 \\
        Pure BERT & 0.296 & 0.892 \\
        \textbf{SemAligner (HumanML3D)} & \textbf{0.554} & \textbf{0.589} \\
        \textit{SemAligner (BABEL)} & \textit{0.594} & \textit{0.515} \\
        \bottomrule
    \end{tabular}
\end{table}

\subsection{Ablation Study on SemAligner: The Role of Contextual Relevance}
\label{appendix:semaligner}

The SemAligner module is pivotal for retrieving high-quality in-context demonstrations, which serve as ``semantic anchors'' for the LLM. To quantify its contribution, we utilized the BABEL dataset to benchmark our fine-tuned SemAligner against two baseline retrieval strategies:
\begin{enumerate}
    \item \textbf{Random Selection:} A baseline representing zero prior knowledge.
    \item \textbf{Original BERT:} A pre-trained, frozen BERT model that selects examples based on generic textual similarity, lacking motion-specific alignment.
\end{enumerate}

As shown in Table~\ref{tab:semaligner_ablation}, \textbf{Pure BERT} barely outperforms \textbf{Random Selection} (0.296 vs. 0.271), confirming that generic embeddings lack the kinematic alignment necessary for motion tasks. In contrast, our \textbf{SemAligner} achieves a substantial performance leap (R-Precision: 0.554), demonstrating strong OOD generalization despite disparate training data. The narrow gap ($\Delta \approx 0.04$) compared to the BABEL-trained version effectively provides evidence against severe overfitting under this evaluation setting, validating SemAligner as a robust, cross-dataset semantic anchor.

\subsection{Validity of ActionCLIP: Robustness to Domain Shift}
\label{appendix:actionclip}

We empirically validate the deployment of ActionCLIP~\cite{wang2021actionclip} as a semantic reward model, leveraging its established zero-shot transfer capabilities. To rigorously assess its reliability under the domain shift to rendered SMPL meshes, we conducted a comparative study against the domain-specialized MotionLLM~\cite{chen2025motionllm}. As shown in Table~\ref{tab:oracle_sanity}, ActionCLIP achieves a Top-3 R-Precision of 0.789, which is comparable to that of MotionLLM (0.804). This comparable performance suggests that ActionCLIP maintains reliable semantic alignment in our simulation environment, effectively bridging the domain gap without necessitating specialized retraining.

\begin{table*}[h]
    \caption{Comparison of training and inference efficiency. The training/preparation-time column reports either task-specific training time or one-time offline preparation time. For ICMPG, the reported 4~h corresponds to the one-time offline training of the VQ-VAE and ActionCLIP modules, which are reused across tasks; therefore, ICMPG requires no task-specific retraining. For methods whose inference latency is unavailable or not directly comparable, the latency is marked as ``--''.}
    \label{tab:efficiency_analysis}
    \centering
    \begin{tabular}{l|c|c|c|c}
        \toprule
        Method & Paradigm & Total Training Time & Inference Latency & Simulation Rate (FPS) \\
        \midrule
        T2M-GPT & LLM fine-tuning & 90+ h & $\sim$0.25 s & $\ge$30 FPS \\
        MDM & Diffusion fine-tuning & 70+ h & $\sim$0.6 s & $\ge$30 FPS\\
        MotionGPT (small) & LLM fine-tuning & 130+ h & $\sim$\textbf{0.25 s} & $\ge$30 FPS\\
        AnySkill & RL policy learning & 90+ h & - & $\ge$30 FPS \\
        UniPhys & Diffusion and RL policy learning & 250+ h & - & $\ge$18 FPS \\
        MotionCraft & Zero-shot generation & \textbf{0 h} & $\sim$3 s & $\ge$30 FPS\\
        \midrule
        \textbf{ICMPG} & Inference-time opt. & 4 h & $\sim$10 s & $\ge$30 FPS \\
        \textbf{ICMPG (LoRA)} & LLM fine-tuning & 120+ h & $\sim$0.6 s & $\ge$30 FPS\\
        \bottomrule
    \end{tabular}
\end{table*}

\begin{table}[!t]
    \caption{Validation of Semantic Reward Models on Rendered SMPL Meshes}
    \label{tab:oracle_sanity}
    \centering
    \begin{tabular}{l|cc}
        \toprule
        Model & R-Precision (Top-3) $\uparrow$ & Inference (ms) $\downarrow$ \\
        \midrule
        CLIP (ViT-B/32) & 0.537 & 46 \\
        MotionLLM & \textbf{0.804} & 3285 \\
        \textbf{ActionCLIP (Ours)} & 0.789 & \textbf{42} \\
        \bottomrule
    \end{tabular}
\end{table}

\subsection{Expressiveness of Motion Codebook: Compositional Generalization}
\label{appendix:codebook}

\begin{figure}[h]
    \centering
    \includegraphics[width=\linewidth]{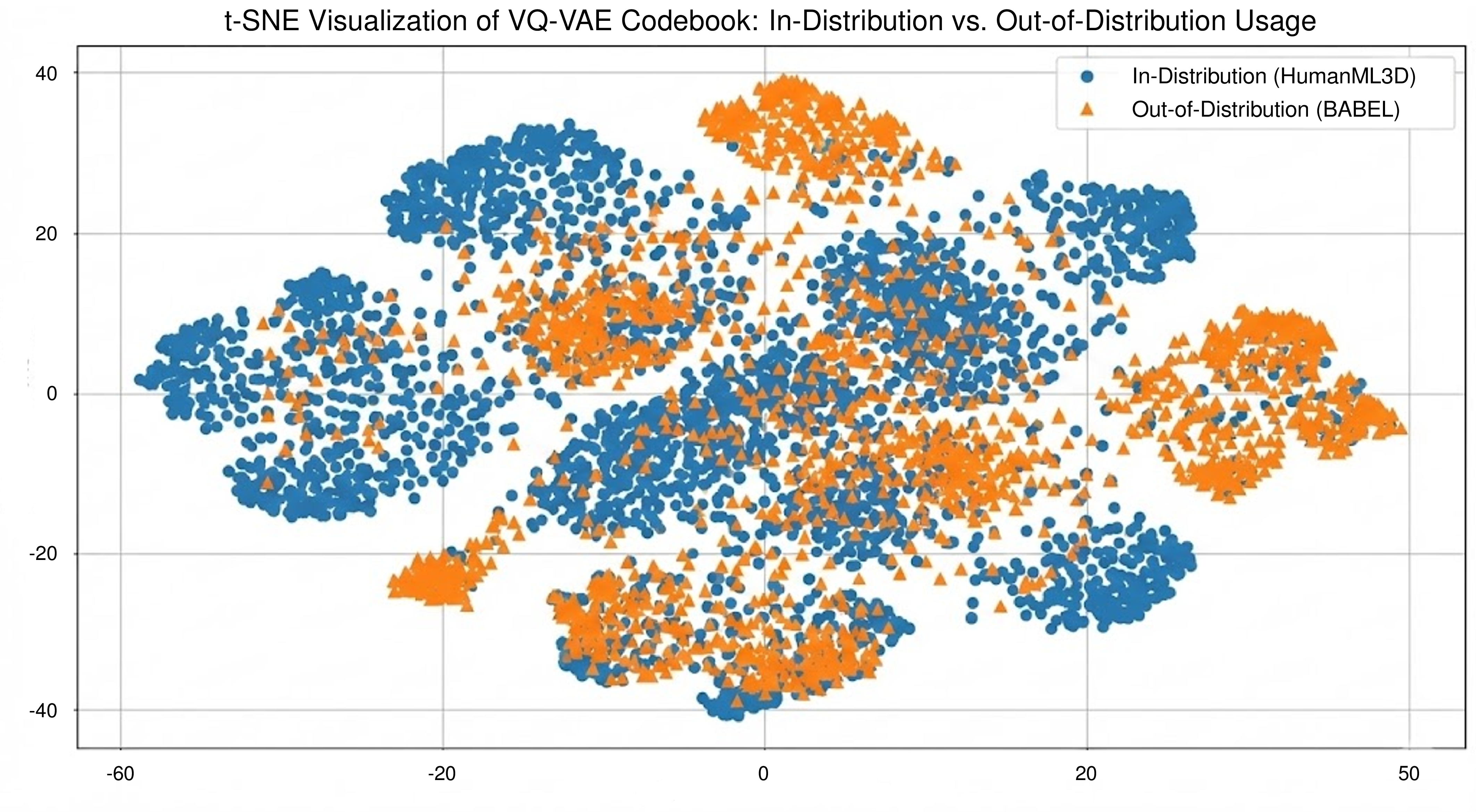}
    \caption{t-SNE visualization of motion codebook utilization. Blue points indicate tokens activated by in-distribution samples from HumanML3D, and orange triangles indicate tokens activated by out-of-distribution samples from BABEL. The substantial overlap suggests that out-of-distribution motions reuse fundamental motion primitives learned from the training data through novel compositions of shared tokens.}
    \label{fig:codebook_tsne}
\end{figure}

To address the concern regarding the representational capacity of a fixed codebook, we posit that open-world generalization is achieved via compositionality, i.e., recombining learned atomic primitives, rather than requiring the creation of new tokens. To validate this, Fig.~\ref{fig:codebook_tsne} presents a t-SNE visualization comparing the codebook utilization of in-distribution (HumanML3D) motions versus out-of-distribution (BABEL) motions.

As illustrated, the OOD tokens (orange) do not form isolated clusters; instead, they are extensively distributed across the latent space and exhibit significant overlap with the in-distribution tokens (blue). This substantial alignment corroborates that ``unseen'' OOD motions are effectively constructed from the same finite set of robust atomic primitives (e.g., steps and bends) captured during training. Consequently, the model's ability to handle novel motions stems from the LLM's capacity to sequence these shared primitives into novel temporal permutations, while the VQ-VAE decoder acts as a continuous interpolator to ensure smooth transitions. This suggests that our codebook possesses sufficient expressiveness for OOD generalization without expanding the vocabulary size.

\subsection{Computational Efficiency: A Strategic Trade-off}
\label{appendix:efficiency}

We present a transparent analysis of the computational profile of ICMPG relative to traditional learning-based baselines. As an inference-time optimization method, ICMPG inherently trades real-time inference speed for zero-shot flexibility. In contrast, the LoRA-trained variant uses a world model and a streamlined pipeline to optimize inference efficiency, ensuring performance that is in line with established training-based approaches.

Table~\ref{tab:efficiency_analysis} details total training time, inference latency, and simulation latency of the deployment in all methods. For some simulation-based or control-based methods, inference latency is unavailable or not directly comparable. As the table shows, LoRA-trained ICMPG achieves training and inference overhead comparable to other LLM-based methods. While the standard inference-time ICMPG incurs higher latency due to physics simulation and closed-loop refinement, we provide the following analysis regarding this:

\begin{itemize}
    \item \textbf{Mitigating the ``Cold Start'' Bottleneck.} Conventional RL-based methods typically lack generalization capabilities for out-of-distribution (OOD) tasks, yielding reference motions that exceed the tracking capabilities of the physics-based policy. Consequently, this necessitates retraining on task-specific datasets, incurring a significant computational overhead of 80--120 hours.
    \item \textbf{Zero-Shot Generalization.} ICMPG synergizes LLM reasoning with a universal physics controller for immediate inference. Achieving ``train-once-for-all'' efficiency comparable to zero-shot MotionCraft, it avoids task-specific retraining and reduces the turnaround time for OOD synthesis.
    \item \textbf{Operational Scope.} The primary objective of ICMPG is not to prioritize real-time interaction but to accelerate the synthesis of high-fidelity, physically plausible motions. The framework features a modular, plug-and-play architecture that is agnostic to the underlying LLM. Crucially, it eliminates the need for task-specific retraining and facilitates rapid deployment across diverse scenarios.
\end{itemize}

\bibliographystyle{IEEEtran}
\bibliography{references}

\begin{IEEEbiography}[{\includegraphics[width=1in,height=1.25in,clip,keepaspectratio]{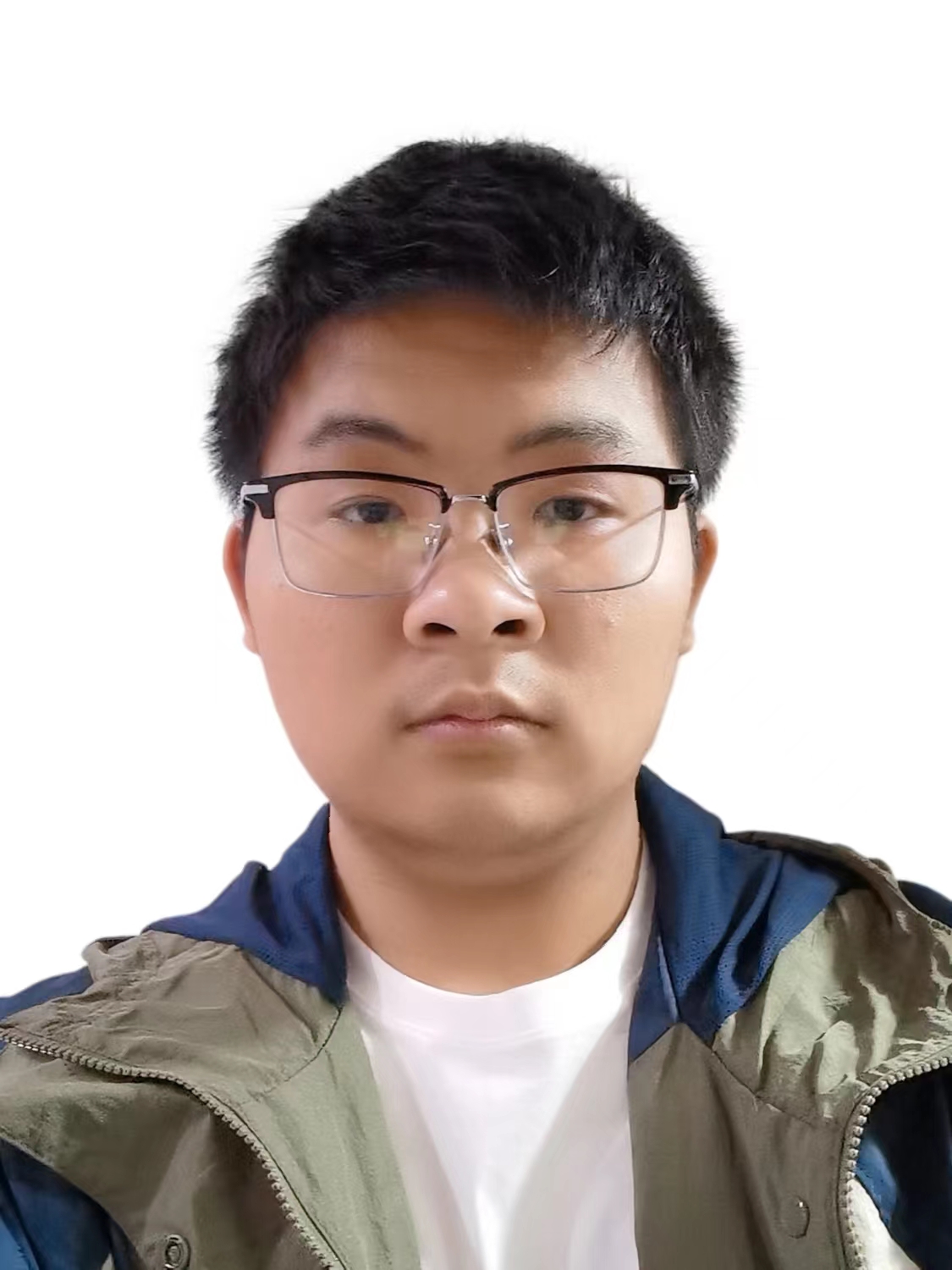}}]{Xiaomeng Fu} received the B.E. degree from Sun Yat-sen University in 2021. He is currently a first-year Ph.D. student at Peng Cheng Laboratory. His research focuses on multimodal generation and AI agents.
\end{IEEEbiography}

\begin{IEEEbiography}[{\includegraphics[width=1in,height=1.25in,clip,keepaspectratio]{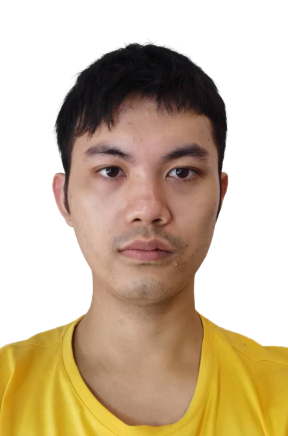}}]{Junfan Lin} received the Ph.D. degree from Sun Yat-sen University. He is currently a postdoctoral researcher at Peng Cheng Laboratory. His research interests include reinforcement learning, embodied AI, and causal inference.
\end{IEEEbiography}

\begin{IEEEbiography}[{\includegraphics[width=1in,height=1.25in,clip,keepaspectratio]{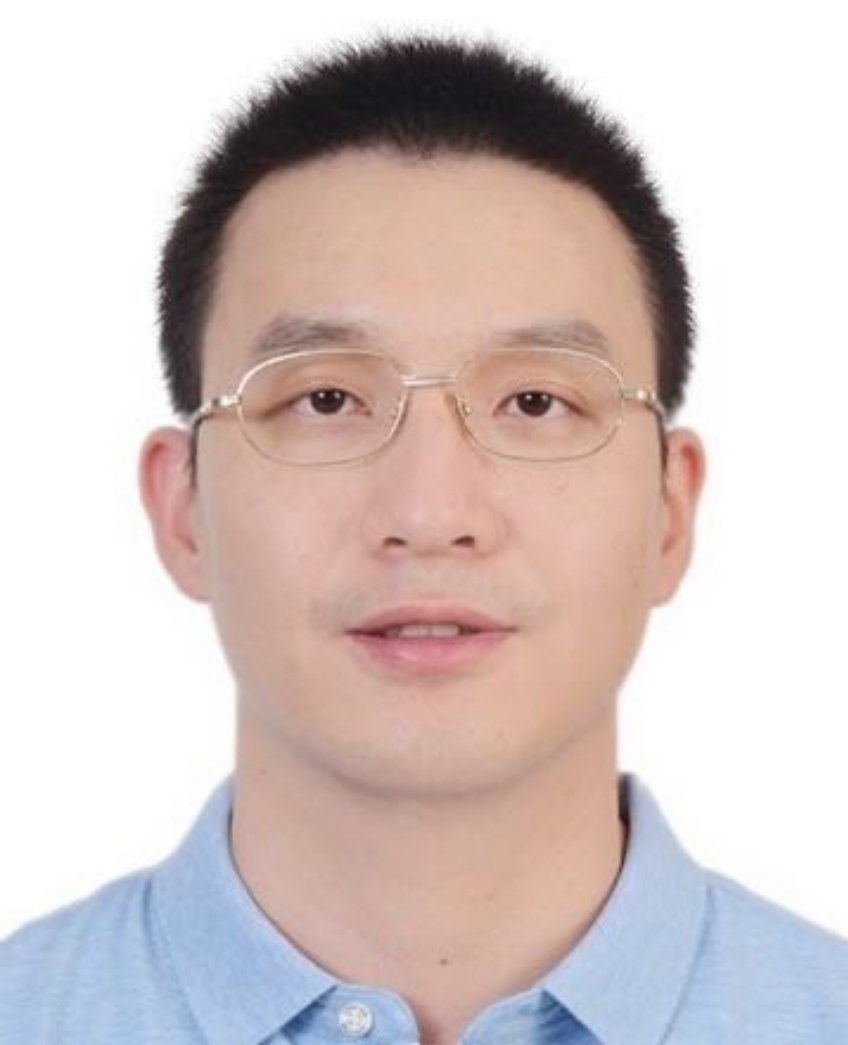}}]{Yang Liu} (M'21) is an Associate Professor at the School of Computer Science and Engineering, Sun Yat-sen University. He received his Ph.D. degree from Xidian University in 2019. His research interests include multimodal reasoning, causality learning, and embodied AI.
\end{IEEEbiography}

\begin{IEEEbiography}[{\includegraphics[width=1in,height=1.25in,clip,keepaspectratio]{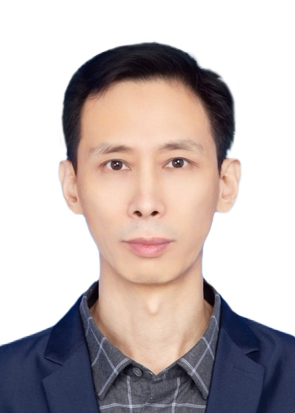}}]{Yaowei Wang} (M'10) is a Full Professor at Harbin Institute of Technology, Shenzhen, and Peng Cheng Laboratory, China. He obtained his Ph.D. degree from Harbin Institute of Technology. His research focuses on computer vision, multimodal perception, and intelligent robotics.
\end{IEEEbiography}

\begin{IEEEbiography}[{\includegraphics[width=1in,height=1.25in,clip,keepaspectratio]{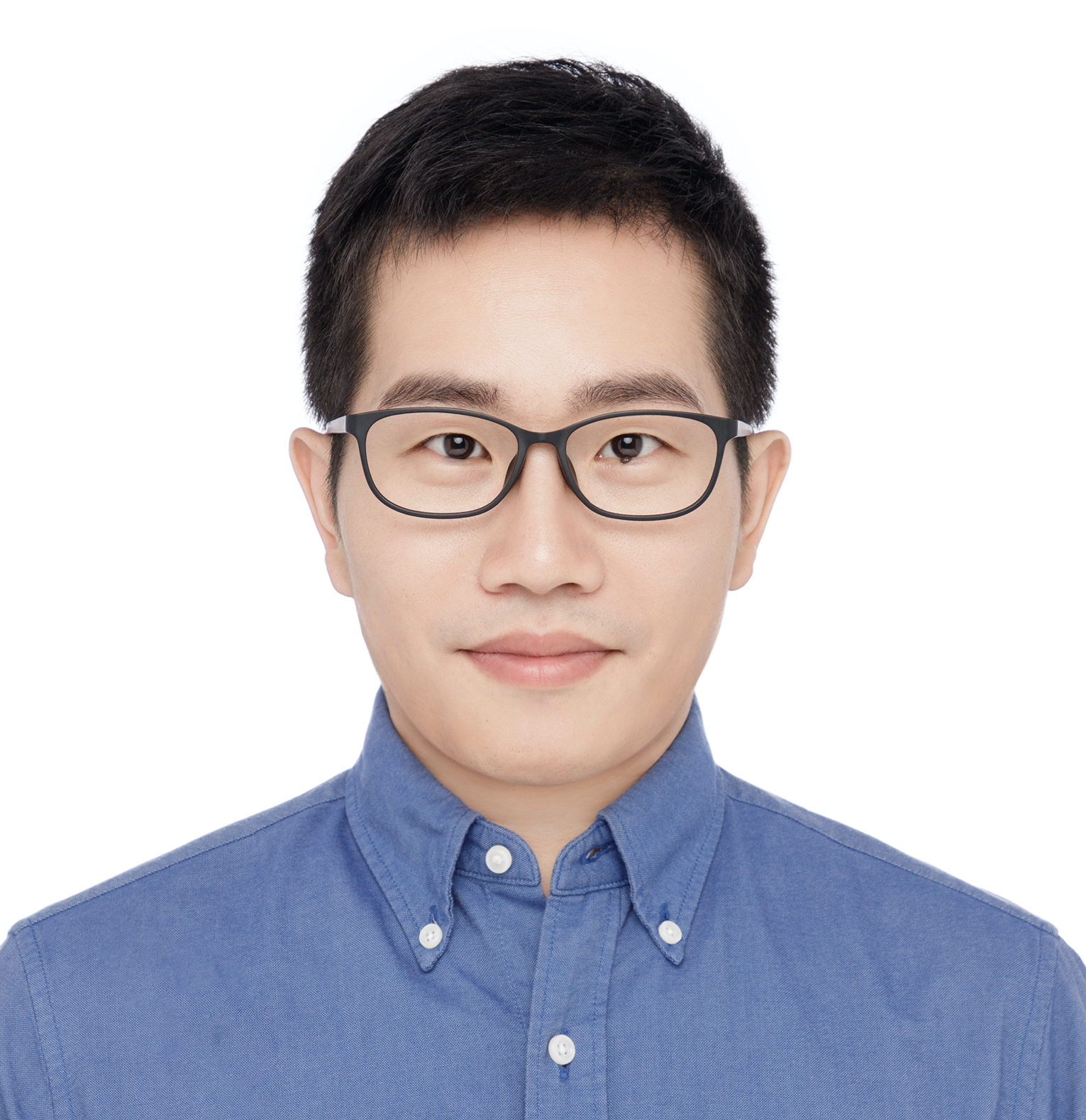}}]{Guanbin Li} (M'15) is currently a full professor in School of Computer Science and Engineering, Sun Yat-sen University. He received his PhD degree from the University of Hong Kong in 2016. His current research interests include computer vision, image processing, and deep learning. He is a recipient of the ICCV 2019 Best Paper Nomination Award. He has authorized and co-authorized on more than 200 papers in top-tier academic journals and conferences. He serves as an area chair for the conference of CVPR and ICCV. He has been serving as a reviewer for numerous academic journals and conferences such as TPAMI, IJCV, TIP, TMM, TCyb, CVPR, ICCV, ECCV and NeurIPS.
\end{IEEEbiography}

\begin{IEEEbiography}[{\includegraphics[width=1in,height=1.25in,clip,keepaspectratio]{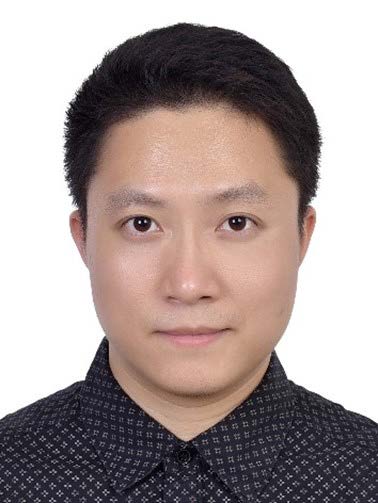}}]{Liang Lin}
(Fellow, IEEE) is a Full Professor of computer science at Sun Yat-sen University. He served as the Executive Director and Distinguished Scientist of SenseTime Group from 2016 to 2018, leading the R\&D teams for cutting-edge technology transferring. He has authored or co-authored more than 300 papers in leading academic journals and conferences, and his papers have been cited by more than 34,000 times. He is an associate editor of IEEE TRANSACTIONS ON NEURAL NETWORKS AND LEARNING SYSTEMS and IEEE TRANSACTIONS
ON MULTIMEDIA, and served as Area Chairs for numerous conferences such as CVPR, ICCV, SIGKDD and AAAI. He is the recipient of numerous awards and honors including Wu Wen-Jun Artificial Intelligence Award, the First Prize of China Society of Image and Graphics, ICCV Best Paper Nomination in 2019, Annual Best Paper Award by Pattern Recognition (Elsevier) in 2018, Best Paper Dimond Award in IEEE ICME 2017, Google Faculty Award in 2012. His supervised PhD students received ACM China Doctoral Dissertation Award, CCF Best Doctoral Dissertation and CAAI Best Doctoral Dissertation. He is a Fellow of IEEE/IAPR. 
\end{IEEEbiography}

\begin{IEEEbiography}[{\includegraphics[width=1in,height=1.25in,clip,keepaspectratio]{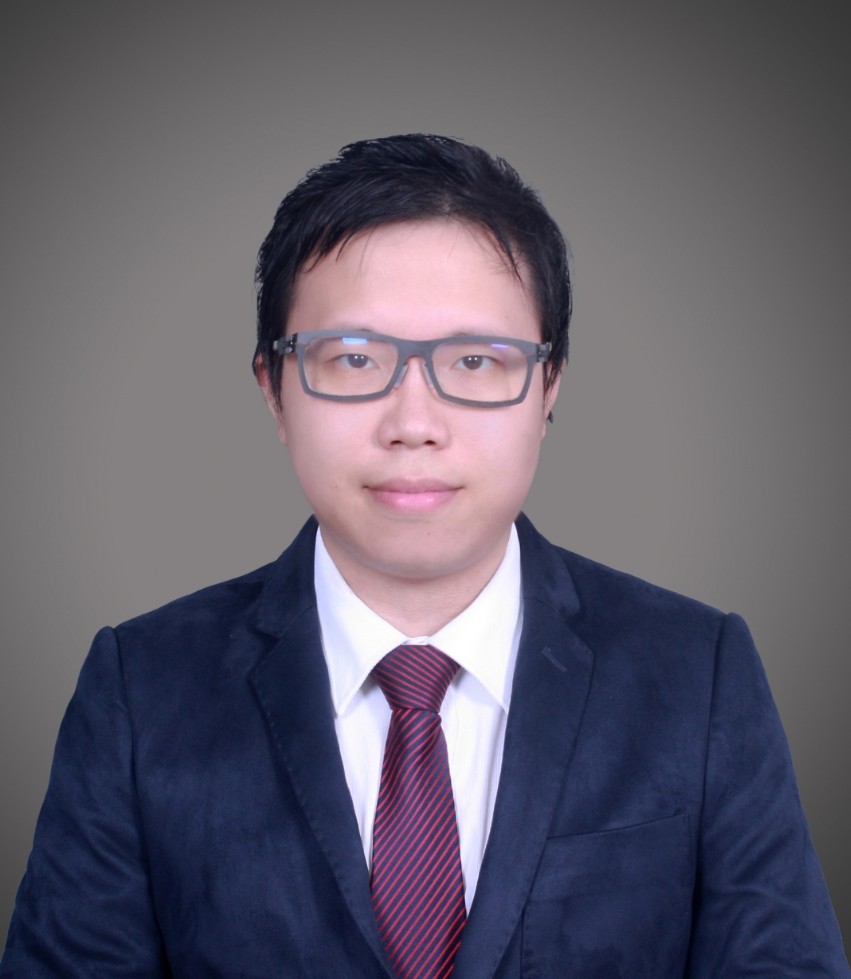}}]{Ziliang Chen} is an Assistant Researcher at Peng Cheng Laboratory. He received his Ph.D. from Sun Yat-sen University. His research interests include large language models and multimodal embodied intelligence. He has published numerous papers in top-tier venues such as ICML, CVPR, and TPAMI.
\end{IEEEbiography}

\end{document}